\documentclass[10pt,twocolumn,letterpaper]{article}

\usepackage[pagenumbers]{cvpr} 

\usepackage{graphicx}
\usepackage{amsmath}
\usepackage{amssymb}
\usepackage{booktabs}

\usepackage{subcaption}
\usepackage{color, colortbl}
\definecolor{Gray}{gray}{0.9}
\usepackage[table,dvipsnames]{xcolor}

\renewcommand*{\ie}{i.e.\@\xspace}
\renewcommand*{\eg}{e.g.\@\xspace}

\usepackage{enumitem}
\setlist{leftmargin=10pt,labelindent=10pt}
\setlist[enumerate]{wide=0pt, leftmargin=10pt, labelwidth=10pt, align=left}

\usepackage[pagebackref,breaklinks,colorlinks]{hyperref}

\usepackage[capitalize]{cleveref}
\crefname{section}{Sec.}{Secs.}
\Crefname{section}{Section}{Sections}
\Crefname{table}{Table}{Tables}
\crefname{table}{Tab.}{Tabs.}

\def\cvprPaperID{9976} 
\def\confName{CVPR}
\def\confYear{2022}

\begin{document}

\title{When Does Contrastive Visual Representation Learning Work?}

\author{
Elijah Cole$^1$ \quad Xuan Yang$^2$ \quad Kimberly Wilber$^2$ \quad Oisin Mac Aodha$^{3, 4}$ \quad Serge Belongie$^{5}$ \\
{\small $^1$Caltech \quad $^2$Google \quad $^3$University of Edinburgh \quad $^4$Alan Turing Institute \quad $^{5}$University of Copenhagen}
}

\maketitle

\begin{abstract}
    Recent self-supervised representation learning techniques have largely closed the gap between supervised and unsupervised learning on ImageNet classification. 
    While the particulars of pretraining on ImageNet are now relatively well understood, the field still lacks widely accepted best practices for replicating this success on other datasets. 
    As a first step in this direction, we study contrastive self-supervised learning on four diverse large-scale datasets.
    By looking through the lenses of data quantity, data domain, data quality, and task granularity, we provide new insights into the necessary conditions for successful self-supervised learning. 
    Our key findings include observations such as: (i) the benefit of additional pretraining data beyond 500k images is modest, (ii) adding pretraining images from another domain does not lead to more general representations, (iii) corrupted pretraining images have a disparate impact on supervised and self-supervised pretraining, and (iv) contrastive learning lags far behind supervised learning on fine-grained visual classification tasks. 
\end{abstract}

\section{Introduction}

Self-supervised learning (SSL) techniques can now produce visual representations which are competitive with representations generated by fully supervised networks for many downstream tasks~\cite{ericsson2020well}. 
This is an important milestone for computer vision, as removing the need for large amounts of labels at training time has the potential to scale up our ability to address challenges in domains where supervision is currently too difficult or costly to obtain. However, with some limited exceptions, the vast majority of current state-of-the-art approaches are developed and evaluated on standard datasets like ImageNet~\cite{russakovsky2015imagenet}. As a result, we do not have a good understanding of how well these methods work when they are applied to other datasets. 

\begin{figure}[t]
    \centering
    \includegraphics[width=0.42\textwidth]{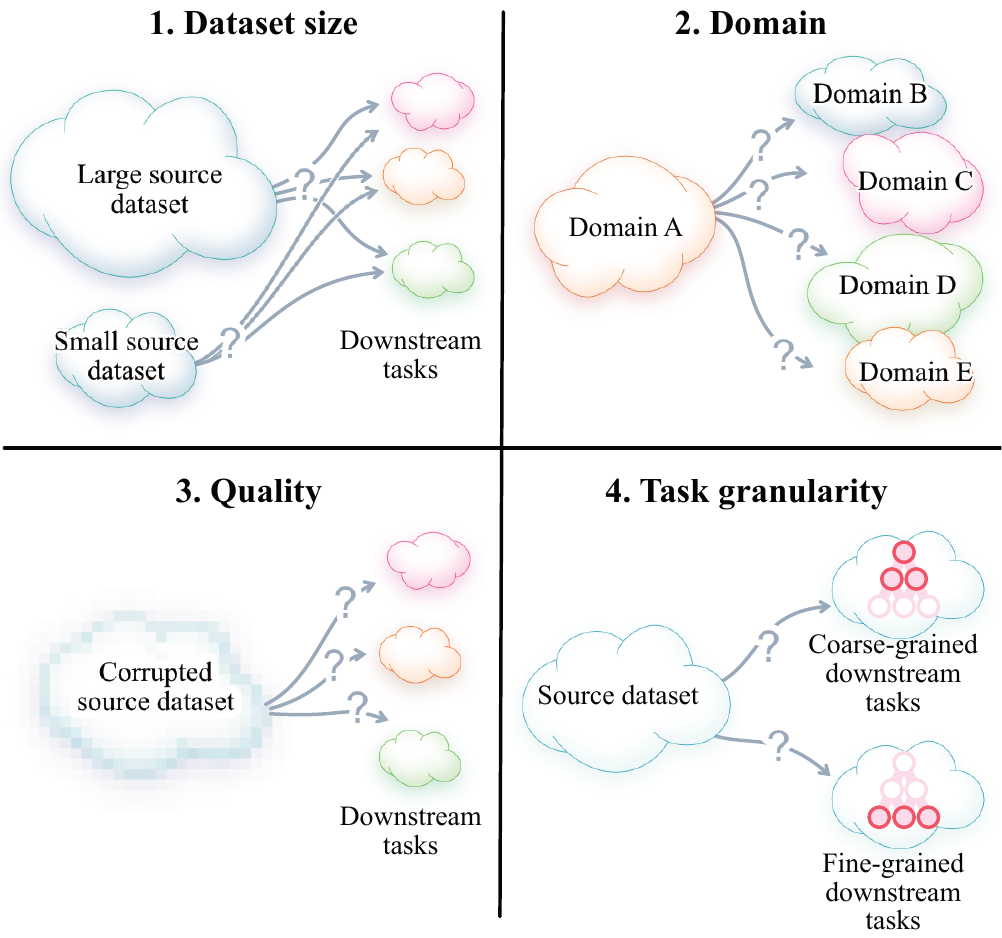}
    \vspace{-6pt}
    \caption{\label{fig:splash}
    \textbf{What conditions are necessary for successful self-supervised pretraining on domains beyond ImageNet?}
    We investigate the impact of self-supervised and supervised training \emph{dataset size}, the downstream \emph{domain}, \emph{image quality}, and the \emph{granularity} of downstream classification tasks. 
    }
    \vspace{-15pt}
\end{figure}

\emph{Under what conditions do self-supervised contrastive representation learning methods produce ``good" visual representations?} 
This is an important question for computer vision researchers because it adds to our understanding of SSL and highlights opportunities for new methods. This is also an important question for domain experts with limited resources who might be interested in applying SSL to real-world problems.
With these objectives in mind, we attempt to answer the following questions:

\noindent\textbf{(i) What is the impact of data quantity?} 
How many unlabeled images do we need for pretraining, and when is it worthwhile to get more?
How much labeled data do we need for linear classifier training or end-to-end fine-tuning on a downstream task? 
In which regimes do self-supervised features rival those learned from full supervision? 

\noindent\textbf{(ii) What is the impact of the pretraining domain?}
How well do self-supervised representations trained on one domain transfer to another? 
Can we learn more general representations by combining datasets? 
Do different pretraining datasets lead to complementary representations?

\noindent\textbf{(iii) What is the impact of data quality?} 
How robust are self-supervised methods to training time image corruption such as reduced resolution, compression artifacts, or noise? Does pretraining on corrupted images lead to poor downstream performance on uncorrupted images?

\noindent\textbf{(iv) What is the impact of task granularity?} Does SSL result in features that are only effective for ``easy" classification tasks, or are they also useful for more challenging, ``fine-grained" visual concepts?

We address the above questions through extensive quantitative evaluation across four diverse large-scale visual datasets (see Figure~\ref{fig:splash}). 
We make several interesting observations and recommendations including: 
\vspace{-8pt}
\begin{itemize}
    \setlength\itemsep{-4pt} 
    \item For an ImageNet-scale dataset, decreasing the amount of unlabeled training data by half (from 1M to 500k images) only degrades downstream classification performance by 1-2\% (Figure~\ref{fig:performance_vs_num_images}). 
    In many contexts this trade-off is reasonable, allowing for faster and cheaper pretraining. This also indicates that current self-supervised methods coupled with standard architectures may be unable to take advantage of very large pretraining sets. 
    
    \item Self-supervised representations that are learned from images from the same domain as the test domain are much more effective than those learned from different domains (Table~\ref{tab:large_dataset_grid_simclr}). 
    Self-supervised training on our current datasets may not be sufficient to learn representations that readily generalize to many contexts. 
    
    \item Neither (i) combining datasets before pretraining (Table~\ref{tab:pooled_data}) nor (ii) combining self-supervised features learned from different datasets (Table~\ref{tab:complementary_representations}) leads to significant performance improvements. 
    More work may be required before self-supervised techniques can learn highly generalizable representations from large and diverse datasets. 
    
    \item Pretraining on corrupted images affects supervised and self-supervised learning very differently (Figure~\ref{fig:image_corruption}). For instance, self-supervised representations are surprisingly sensitive to image resolution. 
    
    \item Current self-supervised methods learn representations that can easily disambiguate coarse-grained visual concepts like those in ImageNet.
    However, as the granularity of the concepts becomes finer, self-supervised performance lags further behind supervised baselines (Figure~\ref{fig:performance_granularity}).
    The contrastive loss may lead to coarse-grained features which are insufficient for fine-grained tasks. 

\end{itemize}

\section{Related Work}
\noindent
\textbf{SSL for visual representations.}
Early self-supervised representation learning methods typically centered around solving hand-designed ``pretext tasks" like patch location prediction~\cite{Doersch2015-vj},
rotation prediction~\cite{gidaris2018unsupervised}, inpainting~\cite{pathak2016context}, cross-channel reconstruction~\cite{zhang2017split}, sorting sequences of video frames~\cite{lee2017unsupervised}, solving jigsaw puzzles \cite{noroozi2016unsupervised}, or colorization~\cite{zhang2016colorful}. 
However, more recent work has explored \emph{contrastive learning-based} approaches where the pretext task is to distinguish matching and non-matching pairs of augmented input images~\cite{henaff2020dataefficient, oord2019representation, tian2019contrastive}.
The prototypical example is SimCLR~\cite{chen2020simple, chen2020big}, which is trained to identify the matching image using a cross-entropy loss.
Other variations on the contrastive SSL framework include
using a momentum encoder to provide large numbers of negative pairs (MoCo)~\cite{he2020momentum,chen2020improved}, 
adaptively scaling the margin in MoCo (EqCo)~\cite{zhu2020eqco},
and contrasting clustering assignments instead of augmented pairs (SwAV)~\cite{caron2020unsupervised}.
Moving beyond the contrastive loss entirely, some papers
recast the problem in a ``learning-to-rank'' framework
(S2R2)~\cite{varamesh2020self}, use simple feature prediction (SimSiam)~\cite{chen2020exploring}, or predict the output of an exponential moving average network (BYOL)~\cite{grill2020bootstrap}.
\cite{frankle2020all} investigates the role of negatives in contrastive learning, though we note that BYOL and SimSiam avoid using negatives explicitly.
In this work, our focus is on self-supervised visual classification. 
We do not explore alternative settings such as supervised contrastive learning~\cite{khosla2020supervised}, contrastive learning in non-vision areas like language~\cite{rethmeier2020long} or audio~\cite{saeed2020contrastive}, or other methods that aim to reduce the annotation burden for representation learning such as large-scale weak supervision~\cite{mahajan2018exploring}. 

\noindent
\textbf{SSL beyond ImageNet.} 
ImageNet classification has long been viewed as the gold standard benchmark task for SSL, and the gap between supervised and self-supervised performance on ImageNet has steadily closed over the last few years~\cite{chen2020simple, he2020momentum, grill2020bootstrap, caron2020unsupervised}.
There is now a growing expectation that SSL should reduce our dependence on manual supervision in challenging and diverse domains which may \emph{not} resemble the traditional object classification setting represented by ImageNet.
A number of papers have studied how well self-supervised representations pretrained on ImageNet perform on downstream tasks like fine-grained species classification~\cite{xiao2020should}, semantic segmentation~\cite{cao2020parametric}, scene understanding~\cite{grill2020bootstrap}, and instance segmentation~\cite{he2020momentum}. 

More recently, researchers have begun to study the effectiveness of contrastive learning when \emph{pretraining} on datasets other than ImageNet. 
In the case of remote sensing, the unique properties of the data have motivated the development of domain-specific contrastive learning techniques~\cite{kang2020deep, ayush2020geography}. 
In the medical domain, where images tend to be very dissimilar to ImageNet, it has been shown that contrastive pretraining on domain-specific images leads to significant gains compared to pretraining on ImageNet~\cite{sowrirajan2020moco, chen2020big}. 
\cite{kotar2021contrasting} compared the representations learned from five different datasets, and showed that in most cases the best performing representations came from pretraining on similar datasets to the downstream task. 
In the case of fine-grained data, \cite{van2021benchmarking} found that contrastive pretraining on images of animals and plants did not lead to superior performance on downstream bird classification compared to pretraining on ImageNet. 
These apparently conflicting observations may be explained by the relationship between the pretraining and downstream data distributions, which we investigate in our experiments.
\cite{zhao2021makes} and \cite{van2021revisiting} pretrained on several different datasets and showed that there was surprisingly little impact on downstream detection and segmentation performance, unless synthetic data was used for pretraining~\cite{zhao2021makes}.  
\cite{tian2021divide} pretrained on very large datasets (JFT-300M~\cite{sun2017revisiting} and YFCC100M~\cite{thomee2016yfcc100m}), but did not observe an improvement over ImageNet pretraining in the standard regime.  

We build on the above analysis by performing controlled, like-for-like, comparisons of SSL on several large datasets. 
This allows us to separate dataset-specific factors from general patterns in SSL performance, and deliver new insights into the necessary conditions for successful pretraining.

\noindent
\textbf{Analysis of SSL.} 
A number of works have explored questions related to the conditions under which SSL is successful. 
\cite{sariyildiz2021concept} showed that self-supervised representations generalize better than supervised ones when the downstream concepts of interest are less semantically similar to the pretraining set. 
\cite{ericsson2020well} showed that contrastive pretraining on ImageNet performs well on downstream tasks related to object recognition in natural images, while leaving more general study of pretraining in different domains to future work. 
While these works show that SSL on ImageNet can be effective, our experiments demonstrate that current SSL methods can perform much worse than supervised baselines on non-ImageNet domains, \eg fine-grained classification. 

Existing work has also investigated other aspects of SSL, \eg
\cite{purushwalkam2020demystifying} examined the invariances learned, \cite{chen2020intriguing} showed that easily learned features can inhibit the learning of more discriminative ones, ~\cite{zhao2021makes,van2021revisiting,chen2020simple} explored the impact of different image augmentations, \cite{chen2020intriguing,van2021revisiting} compared representations from single vs. multi-object images,  and \cite{chen2020simple,Goyal2019-ws} varied the backbone model capacity. 
Most relevant to our work are studies that vary the amount of data in the pretraining dataset, \eg~\cite{yang2020transfer,zhao2021makes,kotar2021contrasting,van2021revisiting}. 
We extend this analysis by presenting a more detailed evaluation of the impact of the size of the unlabeled and labeled datasets, and investigate the role of data quality, data domain, and task granularity. 

\vspace{-3pt}
\section{Methods}
\vspace{-3pt}

\noindent
\textbf{Datasets.} We perform experiments on four complementary large-scale datasets: ImageNet \cite{deng2009imagenet}, iNat21 \cite{van2021revisiting}, Places365 \cite{zhou2017places}, and GLC20 \cite{cole2020geolifeclef}.
Collectively, these datasets span many important visual properties, including: curated vs.~``in-the-wild" images, fine- vs.~coarse-grained categories, and object-centric images vs.~scenes.
Each dataset has at least one million images, which allows us to make fair comparisons against the traditional ImageNet setting. 
ImageNet (1.3M images, 1k classes) and Places365 (1.8M images, 365 classes) are standard computer vision datasets, so we will not describe them in detail. 
For ImageNet, we use the classic ILSVRC2012 subset of the full ImageNet-21k dataset.
For Places365, we use the official variant ``Places365-Standard (small images)" where all images have been resized to 256x256. 
iNat21 (2.7M images, 10k classes) contains images of plant and animal species and GLC20 (1M images, 16 classes) consists of remote sensing images. As both are recent datasets, we discuss them in the supplementary material. 

\noindent
\textbf{Fixed-size subsets.} For some experiments we control for dataset size by creating subsampled versions of each dataset with sizes: 1M, 500k, 250k, 125k, and 50k images.
We carry out this selection only once, and the images are chosen uniformly at random. 
We refer to these datasets using the name of the parent dataset followed by the number of images in parentheses, \eg ImageNet (500k).
Note that subsets of increasing size are \emph{nested}, so \eg ImageNet (500k) includes all of the images in ImageNet (250k). 
These subsets are also \emph{static} across experiments, \eg ImageNet (500k) always refers to the same set of 500k images.
With the exception of Figures~\ref{fig:performance_vs_num_images}~and~\ref{fig:self_sup_method_comparison}, we use the full dataset for any type of supervised training (\ie linear evaluation, fine tuning, or supervised training from scratch). 
We always report results on the same test set for a given dataset, regardless of the training subset used.

\noindent
\textbf{Training details.}
All experiments in this paper are based on a ResNet-50 \cite{he2016deep} backbone, which is standard in the contrastive learning literature \cite{chen2020simple,caron2020unsupervised,he2020momentum}. 
We primarily perform experiments on SimCLR~\cite{chen2020simple}, a simple and popular contrastive learning method that contains all the building blocks for state-of-the-art self-supervised algorithms. 
We follow the standard protocol of first training with self-supervision alone and then evaluating the learned features using linear classifiers or end-to-end fine-tuning.
Unless otherwise specified, we use hyperparameter settings based on \cite{chen2020simple} for all methods and datasets. 
While this may not lead to maximal performance, it is likely to be representative of how these methods are used in practice -- due to the high computational cost of contrastive pretraining, extensive hyperparameter tuning is not feasible for most users. 
We also consider MoCo~\cite{he2020momentum} and BYOL~\cite{grill2020bootstrap} in Figure~\ref{fig:self_sup_method_comparison}.
Full training details are provided in the supplementary material.

\vspace{-3pt}
\section{Experiments}
\vspace{-3pt}
We now describe our experiments in which we investigate the impact of data quantity, data domain, data quality, and task granularity on the success of contrastive learning.

\subsection{Data quantity}\label{sec:data_quantity}

First we consider the question of how much data is required to learn a ``good" representation using SSL.
There are two important notions of data quantity: (i) the number of \emph{unlabeled images} used for pretraining and (ii) the number of \emph{labeled images} used to subsequently train a classifier. 
Since labels are expensive, we would like to learn representations that generalize well with as few labeled images as possible.
While unlabeled images are cheap to acquire, they still incur a cost because pretraining time is proportional to the size of the pretraining set.
To understand when SSL is cost-effective, we need to understand how performance depends on these two notions of data quantity.

\begin{figure*}
    \centering
    \begin{subfigure}[t]{\textwidth}
    \captionsetup{justification=centering}
    \includegraphics[width=0.32\textwidth]{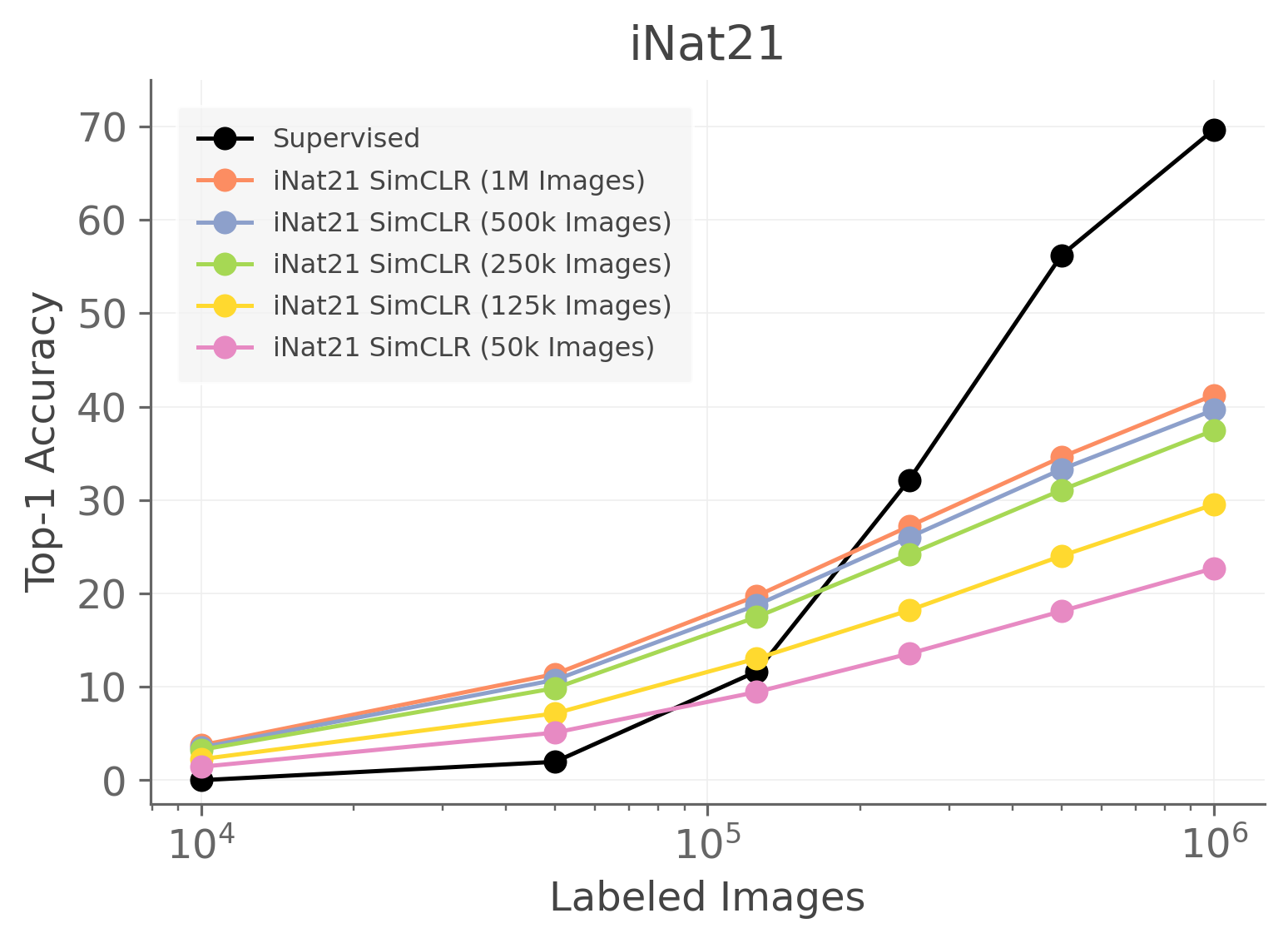}
    \includegraphics[width=0.32\textwidth]{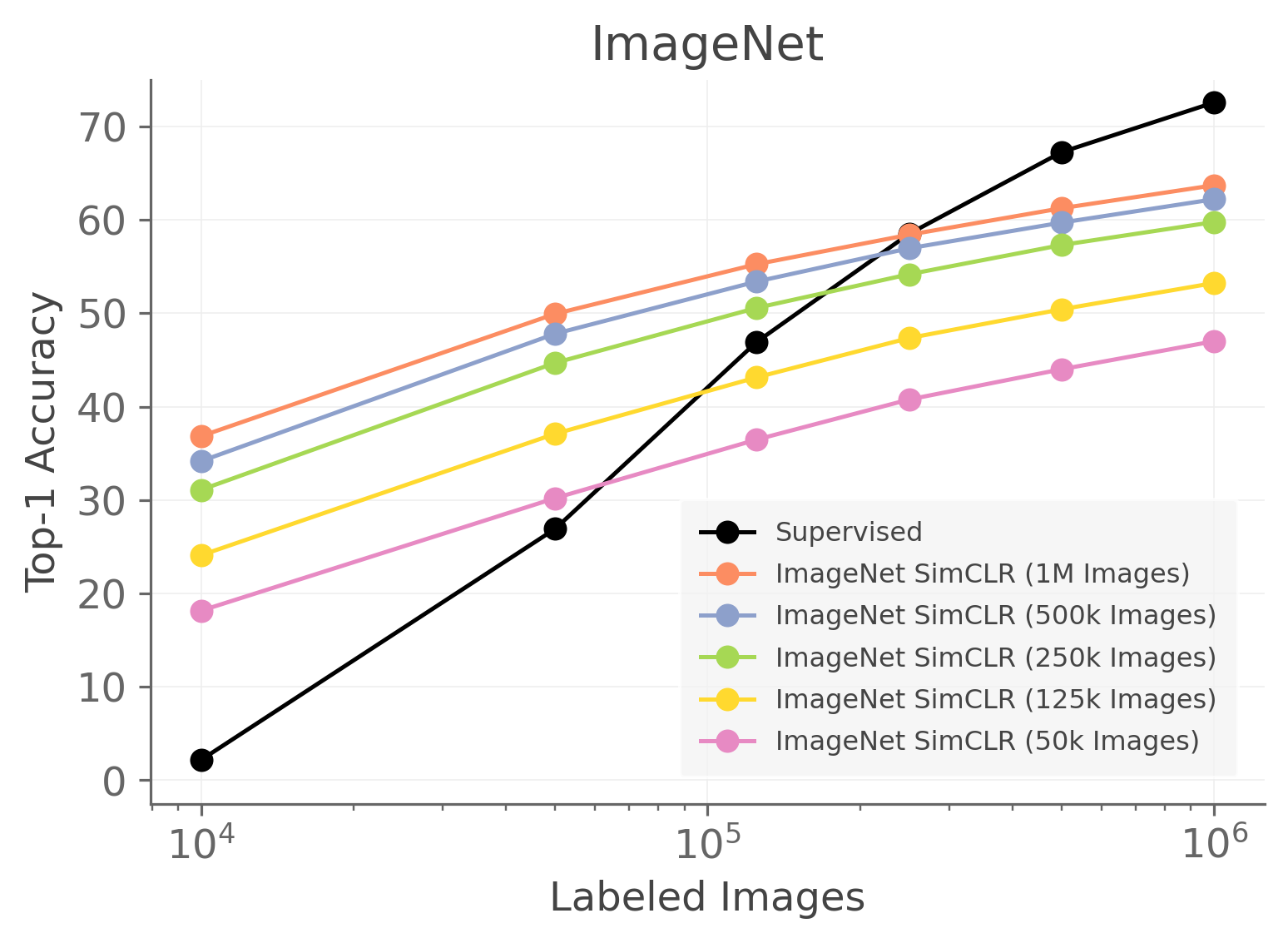}
    \includegraphics[width=0.32\textwidth]{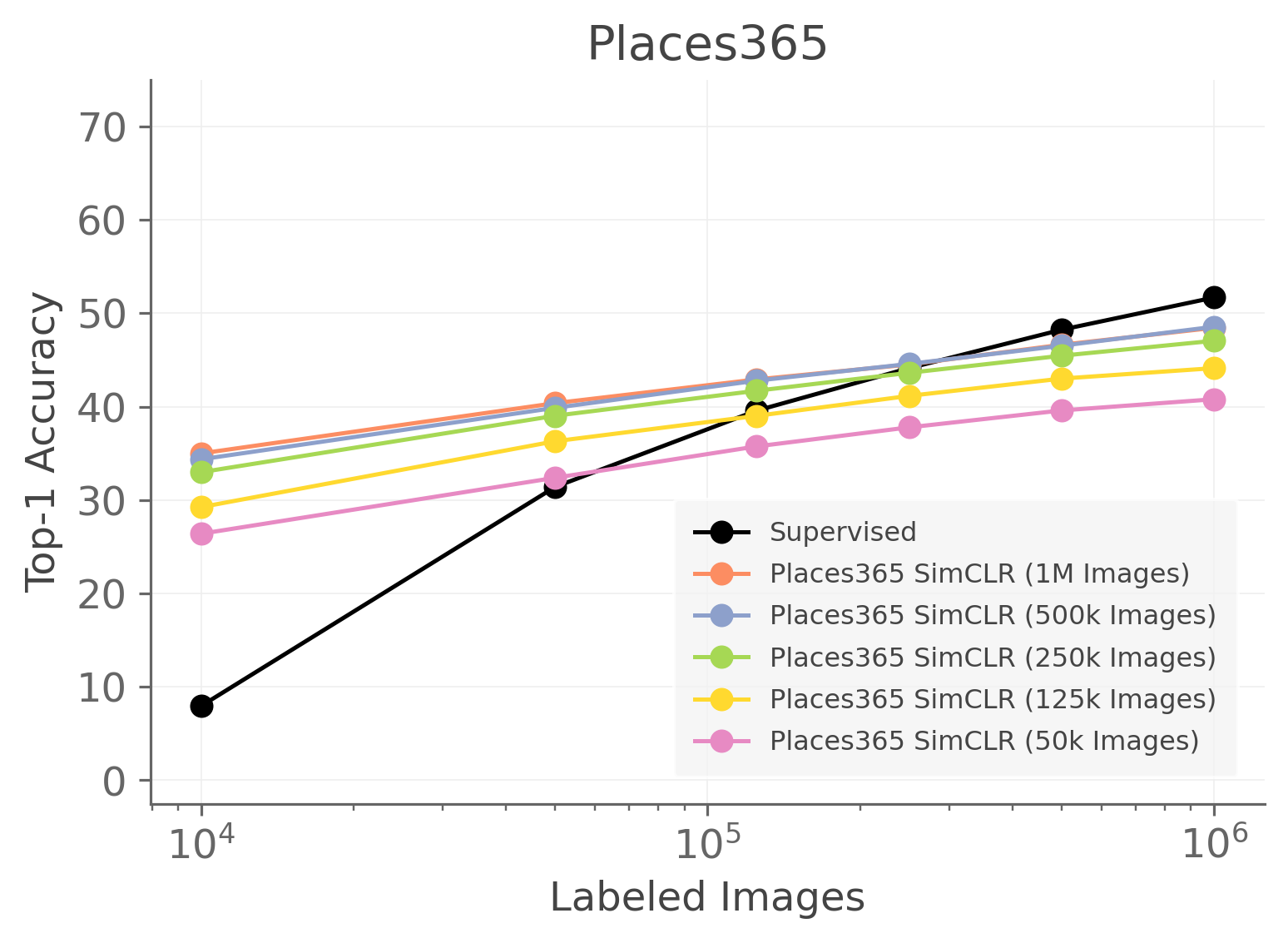}
    \caption{Linear Evaluation}
    \end{subfigure}
    \begin{subfigure}[t]{\textwidth}
    \includegraphics[width=0.32\textwidth]{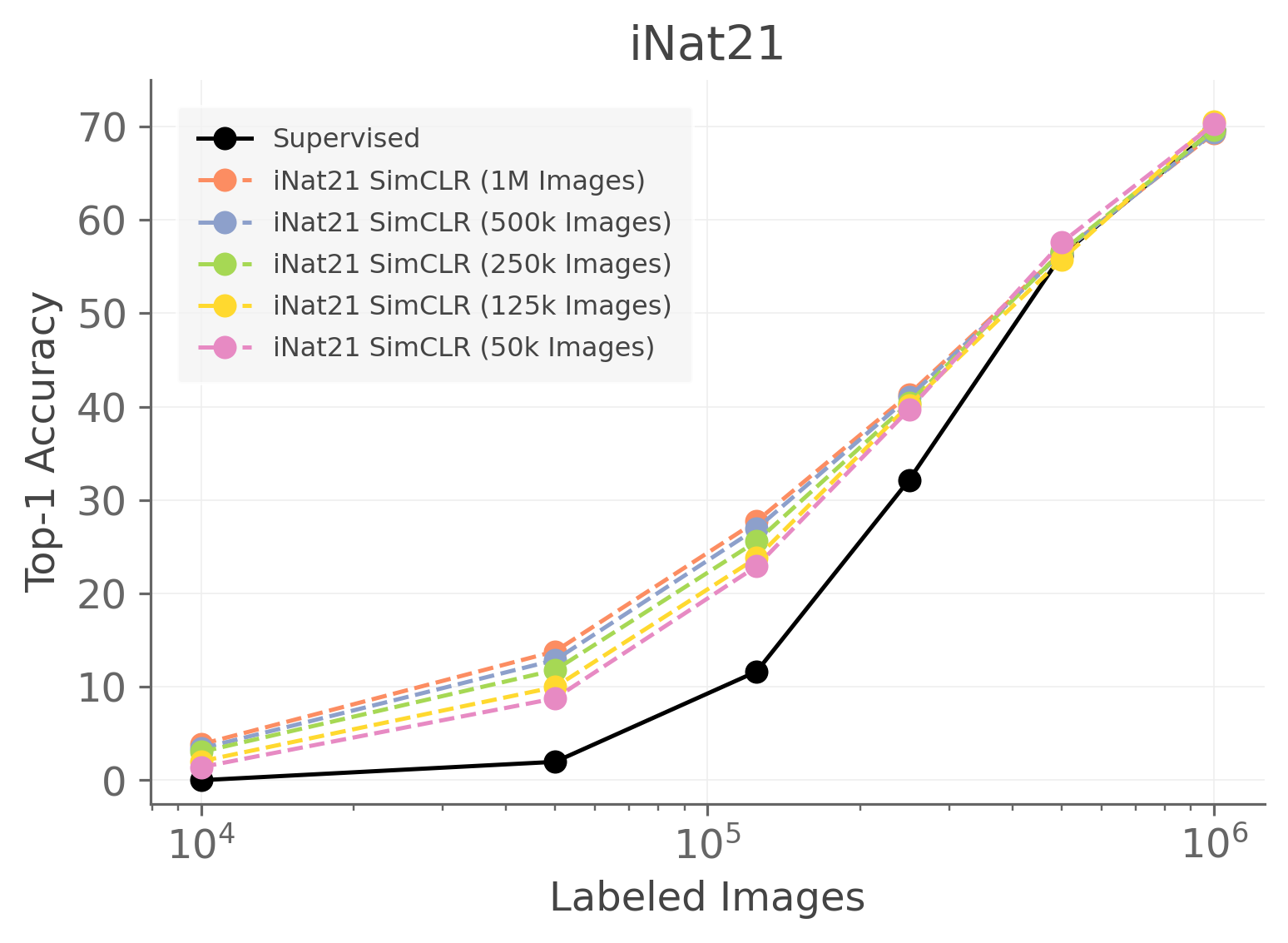}
    \includegraphics[width=0.32\textwidth]{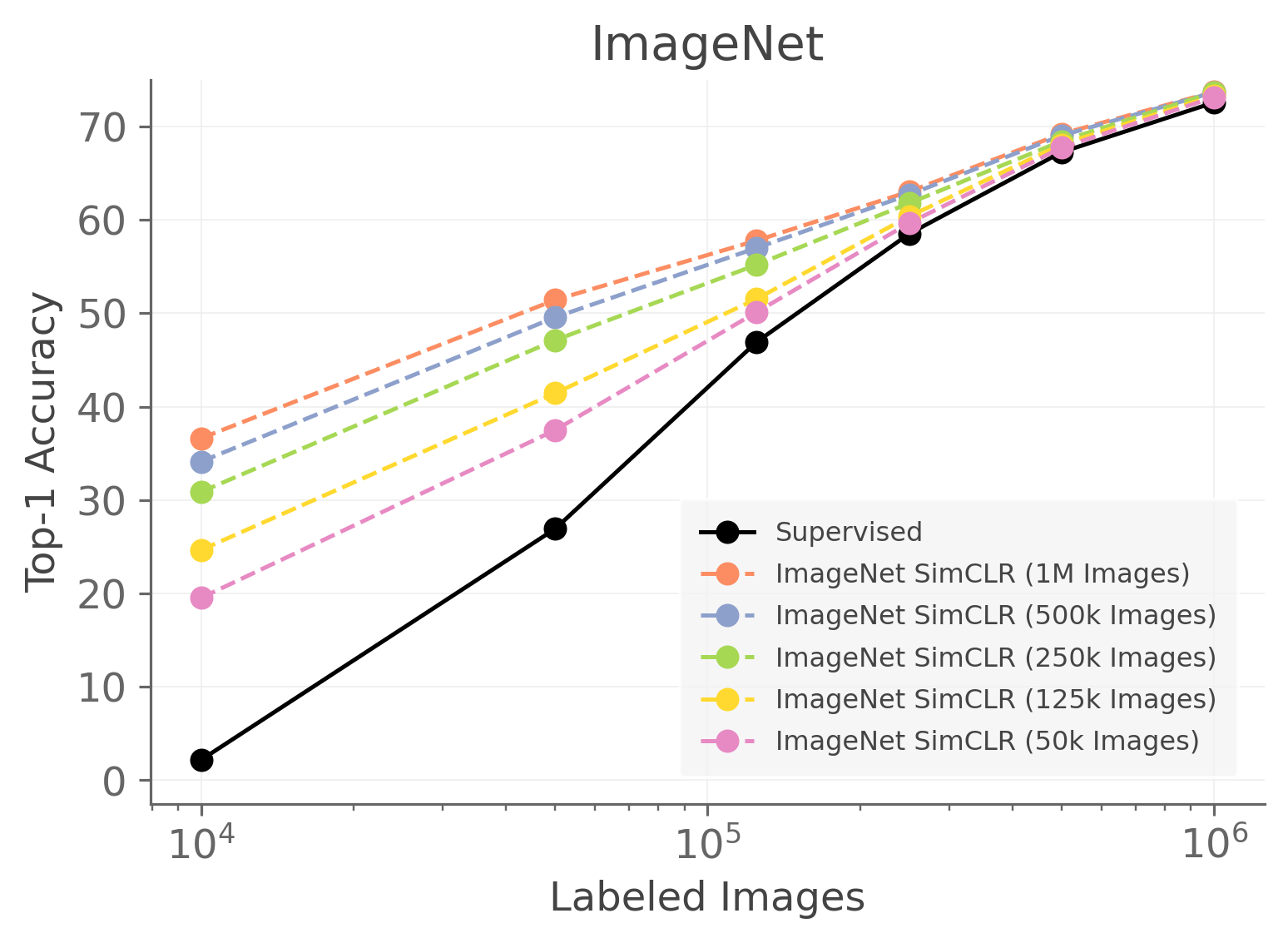}
    \includegraphics[width=0.32\textwidth]{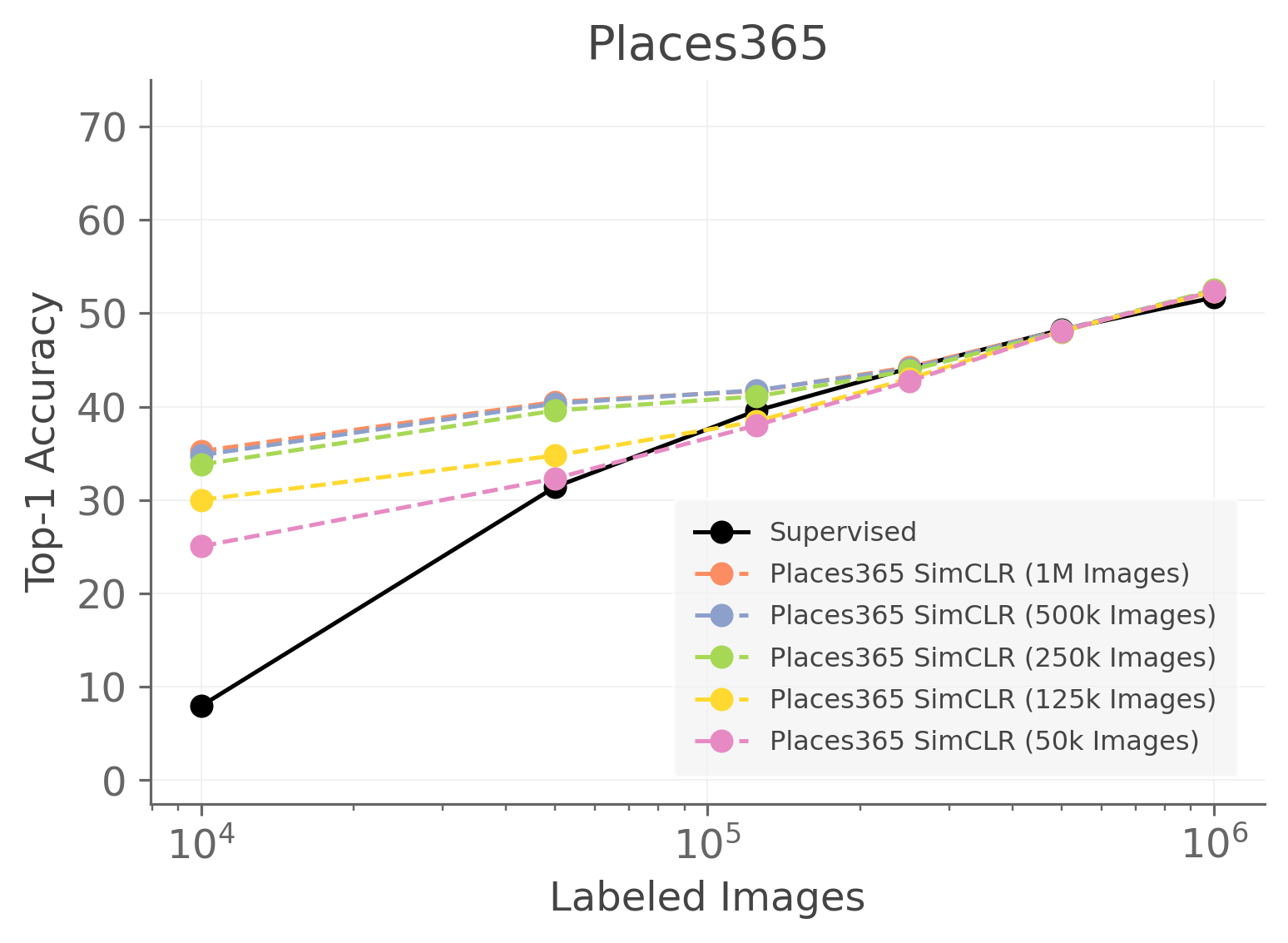}
    \caption{Fine-Tuning}
    \end{subfigure}
    \vspace{-8pt}
    \caption{
    \textbf{How much data does SimCLR need?}
    Linear evaluation results (top row) and fine-tuning results (bottom row) as a function of the number of \emph{unlabeled images} used for pretraining and the number of \emph{labeled images} used for downstream supervised training.
    The ``Supervised" curve (black) corresponds to training from scratch on different numbers of labeled images.
    It is the same for the top and bottom plots in each column.
    Most SSL papers focus on the ``high data" regime, using $\sim 10^6$ images (e.g. all of ImageNet) for both pretraining and classifier supervision, but there are significant opportunities for improvement in the ``low-data" regime.
    Even with $10^6$ labeled images for linear classifier training, SimCLR performs far worse than supervised learning on iNat21, suggesting that iNat21 could be a more useful SSL benchmark than ImageNet in future. 
    }
    \label{fig:performance_vs_num_images}
    \vspace{-10pt}
\end{figure*}

To study this question, we pretrain SimCLR using different numbers of unlabeled images. Each pretrained representation is then evaluated using different numbers of labeled images. 
In Figure~\ref{fig:performance_vs_num_images} we present these results for iNat21 (left column), ImageNet (center column), and Places365 (right column). 
We also include results for supervised training from scratch (in black).
We show linear evaluation results in the top row and corresponding fine-tuned results in the bottom row.
Each curve in a figure corresponds to a different pretrained representation.
The points along a curve correspond to different amounts of supervision used to train a linear classifier or fine-tune the network.

\noindent
\textbf{There is little benefit beyond 500k pretraining images.}
The gap between the 500k (blue) and 1M (orange) pretraining image curves is typically less than 1-2\% in top-1 accuracy.
This means that for a dataset with one million images, we can trade a small decrease in accuracy for a 50\% decrease in pretraining time. 
If a 2-4\% top-1 accuracy drop is acceptable, then the pretraining set size can be reduced by a factor of four (from 1M to 250k).
However, the difference between 50k (pink) pretraining images and 250k (green) pretraining images is substantial for each dataset, often in excess of 10\% top-1 accuracy. 
We conclude that SimCLR seems to saturate well before we get to ImageNet-sized pretraining sets. 
This is consistent with observations from the supervised learning literature, though more images are required to reach saturation \cite{mahajan2018exploring}. 

\noindent
\textbf{Self-supervised pretraining can be a good initializer when there is limited supervision available.}
In the bottom row of Figure~\ref{fig:performance_vs_num_images} we see that when only 10k or 50k labeled images are available, fine-tuning a SimCLR representation is significantly better than training from scratch. 
When supervision is plentiful, fine-tuned SimCLR representations achieve performance similar to supervised training from scratch. 
It is interesting to compare this to findings from the supervised setting which suggest that networks which are initially trained on distorted (i.e. augmented) images are unable to recover when subsequently trained with undistorted ones~\cite{achille2017critical}.

\noindent
\textbf{Self-supervised representations can approach fully supervised performance for some datasets, but only by using lots of labeled images.}
The ultimate goal of SSL is to match supervised performance without the need for large amounts of labeled data.
Suppose we consider the right-most point on the black curves in Figure~\ref{fig:performance_vs_num_images} as a proxy for ``good" supervised performance.
Then in both the linear and fine-tuned cases, the gap between SimCLR (pretrained on 1M images) and ``good" supervised performance is quite large unless well over 100k labeled images are used. 
For instance, the gap between ``good" supervised performance and a classifier trained using 50k labeled images on top of SimCLR (1M) is around 11\% (11\%) for Places365, 23\% (21\%) for ImageNet, and 58\% (56\%) for iNat21 in the linear (and fine-tuned) case. 
Although SSL works well when lots of supervision is available, further innovation is needed to improve the utility of self-supervised representations in the low-to-moderate supervision regime. 

\begin{figure}
    \centering
    \includegraphics[width=0.36\textwidth]{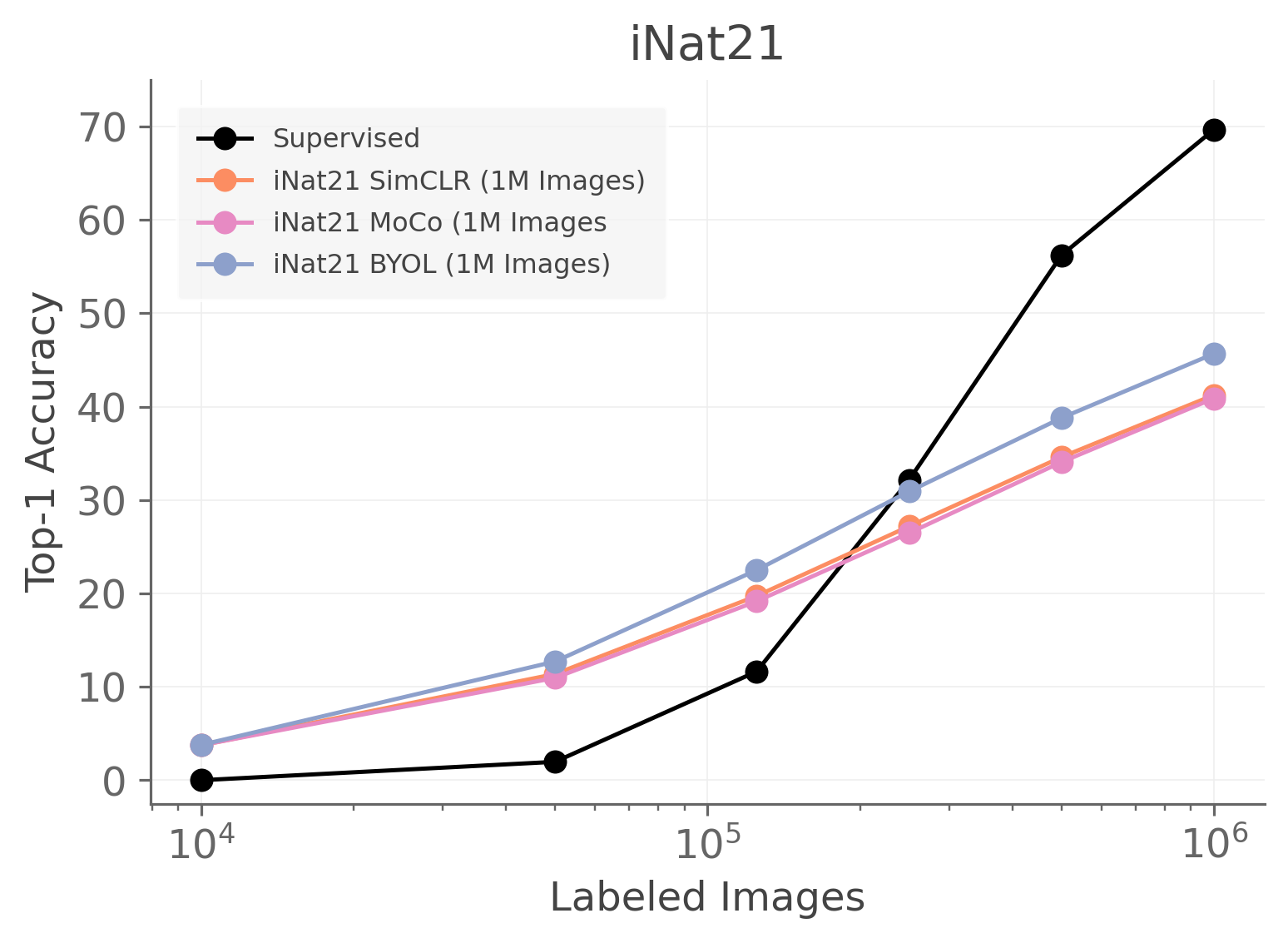}
    \vspace{-8pt}
    \caption{
    \textbf{How does SimCLR compare to other self-supervised methods?}
    Linear evaluation results on iNat21 for SimCLR, MoCo, and BYOL. All methods are pretrained on 1M images for 1000 epochs and follow the same linear evaluation protocol. The more recent BYOL performs better than the others, but a large gap remains to supervised performance. 
    }
    \vspace{-12pt}
    \label{fig:self_sup_method_comparison}
\end{figure}

\noindent
\textbf{iNat21 is a valuable SSL benchmark.} 
Figure~\ref{fig:performance_vs_num_images} shows a surprisingly large gap ($\sim 30\%$) between supervised and self-supervised performance on iNat21 in the high supervision regime. In Figure~\ref{fig:self_sup_method_comparison} we see that other SSL methods exhibit similar limitations. The newer BYOL outperforms MoCo and SimCLR, but a considerable gap ($\sim 25\%$) remains. The high supervised performance shows that the task is possible, yet the self-supervised performance remains low. It seems that iNat21 reveals challenges for SSL that are not apparent in ImageNet, and we believe it is a valuable benchmark for future SSL research. 

\subsection{Data domain}\label{sec:data_domain}

In the previous section we observed that increasing the pretraining set size yields rapidly diminishing returns. 
In this section we consider a different design choice: \emph{what kind of images} should we use for pretraining?
Since most contrastive learning papers only pretrain on ImageNet, this question has not received much attention. We take an initial step towards an answer by studying the properties of SimCLR representations derived from four pretraining sets drawn from different domains.

We train SimCLR on iNat21 (1M), ImageNet (1M), Places365 (1M), and GLC20 (1M). 
By holding the pretraining set size constant, we aim to isolate the impact of the different visual domains.
We present in-domain and cross-domain linear evaluation results for each representation in Table~\ref{tab:large_dataset_grid_simclr}. 
In Table~\ref{tab:pooled_data} we consider the effect of pretraining on \emph{pooled datasets}, \ie new image collections built by shuffling together existing datasets. 
Finally, in Table~\ref{tab:complementary_representations} we study different \emph{fused representations}, which are formed by concatenating the outputs of different feature extractors. 

\noindent
\textbf{Pretraining domain matters.}
In Table~\ref{tab:large_dataset_grid_simclr} we see that in-domain pretraining (diagonal entries) consistently beats cross-domain pretraining (off-diagonal entries). 
The gap can be surprisingly large, \eg in-domain pretraining provides a 12\% boost on iNat21 compared to the best cross-domain pretraining (ImageNet). 
One might have expected that a visually diverse dataset like ImageNet would lead to a better self-supervised representation than a more homogeneous dataset like GLC20 (even when evaluating on GLC20) but this is not what we observe. 

The off-diagonal entries of Table~\ref{tab:large_dataset_grid_simclr} show that training SimCLR on ImageNet leads to the best cross-domain performance, while GLC20 leads to the worst cross-domain performance. 
Since the pretraining protocols and dataset sizes are held constant, we suggest that the characteristics of the image sets themselves are responsible for the differences we observe.
The strong cross-domain performance of SimCLR pretrained on ImageNet may be due to \emph{semantic similarity} -- perhaps it is better to pretrain on a dataset that is semantically similar to the downstream task, even in a self-supervised context.
This makes sense because there are classes in ImageNet that are similar to classes in iNat21 (animals) and Places365 (scenes). 
This also explains the weak performance of GLC20, since remote sensing imagery is not similar to the other datasets. 

\begin{table}[tb]
\centering
\scriptsize
\begin{tabular}{l|c|c|c|c}
Pretraining & iNat21 & ImageNet & Places365 & GLC20 \\ \hline
iNat21 (1M) SimCLR & \textbf{0.493} & \underline{0.519} & 0.416 & 0.707 \\
ImageNet (1M) SimCLR & \underline{0.373} & \textbf{0.644} & \underline{0.486} & \underline{0.716} \\
Places365 (1M) SimCLR & 0.292 & 0.491 & \textbf{0.501} & 0.693 \\ 
GLC20 (1M) SimCLR & 0.187 & 0.372 & 0.329 & \textbf{0.769} \\ \hline
Supervised (All Images) & 0.791 & 0.741 & 0.539 & 0.826 \\
\end{tabular}
\vspace{-6pt}
\caption{
\textbf{Does pretraining domain matter?}
Linear evaluation results for representations derived from different million-image datasets. 
We train the linear classifiers using the full training sets.
The results in the ``Supervised" row correspond to supervised training from scratch on the full training set. 
We report MAP for GLC20 and top-1 accuracy for other datasets.
In all cases, in-domain pretraining outperforms cross-domain pretraining. 
In each column we highlight the \textbf{best} and \underline{second-best} results.
}
\label{tab:large_dataset_grid_simclr}
\vspace{-12pt}
\end{table}

\noindent
\textbf{Adding cross-domain pretraining data does not necessarily lead to more general representations.} 
We have seen that pretraining on different domains leads to representations with significantly differing capabilities. 
This leads to a natural question: \emph{what happens if we combine our datasets and then learn a representation?}

Table~\ref{tab:pooled_data} gives linear evaluation results for SimCLR pretrained on different ``pooled" datasets.  
In each row, $n$ images from dataset $A$ and $m$ images from dataset $B$ are shuffled together to produce a pretraining set of size $n + m$. 
For instance, the pretraining dataset in the first row of Table~\ref{tab:pooled_data} consists of 250k iNat21 images and 250k ImageNet images shuffled together.

If we compare the ``In-Domain (500k)" row against the (equally sized) pooled datasets in the first three rows of Table~\ref{tab:pooled_data}, we see that the in-domain pretraining on 500k images is always better.
Similarly, the ``In-Domain (1M)" row beats the 1M-image pooled dataset (consisting of 250k images from the four datasets). 
The more diverse pooled pretraining sets always lead to worse performance compared to the more homogeneous pretraining sets of the same size. 

Table~\ref{tab:pooled_data} also allows us to say whether it is worthwhile to \emph{add} pretraining data from a different domain (as opposed to swapping out some in-domain data for some data from a different domain, as we have been discussing so far).
The ``In-Domain (250k)" row is better than the 1M-image pooled dataset and almost all of the 500k-image pooled datasets.
It seems that adding pretraining data from a different domain typically \emph{hurts} performance.
In contrast, Figure~\ref{fig:performance_vs_num_images} shows that increasing the amount of \emph{in-domain} pretraining data consistently improves performance.

We hypothesize that the reason for this lackluster performance is that diverse images are easier to tell apart, which makes the contrastive pretext task easier. 
If the contrastive task is too easy, the quality of the representation  suffers~\cite{frankle2020all,chen2020intriguing}. 
While more investigation is needed, the fact that increasing pretraining data diversity can hurt performance suggests a ``diversity-difficulty trade-off" that should be considered when creating pretraining sets for SSL. 

\begin{table}[tb]
\centering
\scriptsize
\begin{tabular}{c|c|c|c|c|c|c}
\multicolumn{4}{c|}{Pretraining} & \multicolumn{3}{c}{Evaluation}\\ \hline
\rotatebox[origin=c]{70}{iNat21} & \rotatebox[origin=c]{70}{ImageNet} & \rotatebox[origin=c]{70}{Places365} & \rotatebox[origin=c]{70}{GLC20} & \rotatebox[origin=c]{70}{iNat21} & \rotatebox[origin=c]{70}{ImageNet} & \rotatebox[origin=c]{70}{Places365}\\ \hline
250k & 250k & - & - & 0.444 & 0.597 & 0.467\\
- & 250k & 250k & - & 0.334 & 0.596 & 0.490 \\ 
250k & - & 250k & - & 0.428 & 0.531 & 0.483\\
250k & 250k & 250k & 250k & 0.410 & 0.574 & 0.482\\ \hline
\multicolumn{4}{l|}{In-Domain (250k)} & 0.451 & 0.608 & 0.485\\
\multicolumn{4}{l|}{In-Domain (500k)} & 0.477 & 0.629 & 0.499\\
\multicolumn{4}{l|}{In-Domain (1M)} & 0.493 & 0.644 & 0.501\\ \hline
\end{tabular}
\vspace{-6pt}
\caption{
\textbf{The effect of dataset pooling.} Linear evaluation results for self-supervised representations derived from \emph{pooled datasets}, where two or more datasets are shuffled together. 
We train the linear classifiers using the full training sets.
The ``In-Domain" results correspond to pretraining on subsets of the dataset named at the top of the column.
Pooling datasets increases pretraining set size and diversity, but we find that performance \emph{decreases} relative to comparable in-domain pretraining. 
The ``In-Domain (1M)" row corresponds to the diagonal entries of Table~\ref{tab:large_dataset_grid_simclr}.
}
\label{tab:pooled_data}
\vspace{-12pt}
\end{table}

\noindent
\textbf{Self-supervised representations can be largely redundant.} 
From Table~\ref{tab:large_dataset_grid_simclr} it is clear that pretraining on different datasets leads to representations that differ significantly. 
For instance, iNat21 SimCLR beats ImageNet SimCLR on iNat21 (+12.4\% ) and ImageNet SimCLR beats iNat21 SimCLR on ImageNet (+12.7\%). 
Do these representations learn complementary information, or do they just capture the same information to different degrees?

To probe this question we concatenate features from different pretrained networks and carry out linear evaluation on these ``fused" representations.
In Table~\ref{tab:complementary_representations} we present linear evaluation results for fused representations on ImageNet and iNat21.
Combining ImageNet SimCLR and iNat21 SimCLR is worse than ImageNet SimCLR alone on ImageNet (-0.6\%), but better than iNat21 SimCLR alone on iNat21 (+1.4\%). 
These effects are small relative to the $>12\%$ difference between ImageNet SimCLR and iNat21 SimCLR. 
This suggests that the two self-supervised representations are largely redundant. 

There is a larger effect when combining supervised and self-supervised representations.
For iNat21, adding ImageNet Sup. (\ie supervised ImageNet features) on top of iNat21 SimCLR improves performance significantly (+4.7\%). 
However, adding iNat21 Sup.~on top of ImageNet SimCLR actually decreases performance (-4.2\%).
These results are consistent with the hypothesis that dataset semantics are important even for SSL.
Since ImageNet is semantically broader than iNat21 (ImageNet has animal classes, but also many other things), features learned from ImageNet (supervised or self-supervised) should be more helpful for iNat21 than vice-versa.

\begin{table}[tb]
\centering
\footnotesize
\begin{tabular}{l l c | c c}
ImageNet & iNat21 & Dim. & ImageNet & iNat21 \\ \hline
SimCLR & - & 2048 & 0.647 & 0.380 \\ 
- & SimCLR & 2048 & 0.520 & 0.506 \\ 
Sup. & - & 2048 & 0.711 & 0.434  \\ 
- & Sup. & 2048 & 0.490 & \underline{0.769} \\ \hline 
Sup. & Sup. & 4096 & 0.712 & \textbf{0.772} \\ 
SimCLR & SimCLR & 4096 & 0.641 & 0.520 \\ 
SimCLR \& Sup. & - & 4096 & \textbf{0.720} & 0.472 \\ 
-& SimCLR \& Sup. & 4096 & 0.527 & \textbf{0.772} \\ 
SimCLR & Sup. & 4096 & 0.605 & \underline{0.769} \\ 
Sup. & SimCLR & 4096 & \underline{0.717} & 0.553 \\ 
\hline
\end{tabular}
\vspace{-6pt}
\caption{
\textbf{The effect of representation fusion.}
Linear evaluation results for different combinations of supervised and self-supervised representations on ImageNet and iNat21.
We train the linear classifiers using the full training sets.
For comparability, the in-domain supervised results in this table (ImageNet Sup. evaluated on ImageNet and iNat21 Sup. evaluated on iNat21) are for linear classifiers trained on representations learned from full supervision.
``Dim." is the representation dimensionality.
In each column we highlight the \textbf{best} and \underline{second-best} results.
}
\label{tab:complementary_representations}
\vspace{-12pt}
\end{table}

\subsection{Data quality}\label{sec:data_quality}
We have seen that the characteristics of the pretraining data can have a significant impact on the quality of self-supervised representations. 
In this section we dig deeper into this question by studying the impact of pretraining on artificially degraded images. 
This serves two purposes. 
First, this is a practical question since there are many settings where image quality issues are pervasive \eg medical imaging \cite{tang2021data} or camera trap data \cite{beery2018recognition}.
Second, it can help us understand the robustness properties of SSL.

To create a corrupted dataset we apply a particular image corruption to each image in the dataset.
This is a one-time offline preprocessing step, so corruptions that have a random component are realized only once per image. 
Given a corrupted dataset we then pretrain as normal. During linear evaluation, we use the original clean images for training and testing, \ie the corrupted images are only used for pretraining.

\begin{figure}
    \centering
    \includegraphics[width=0.48\textwidth]{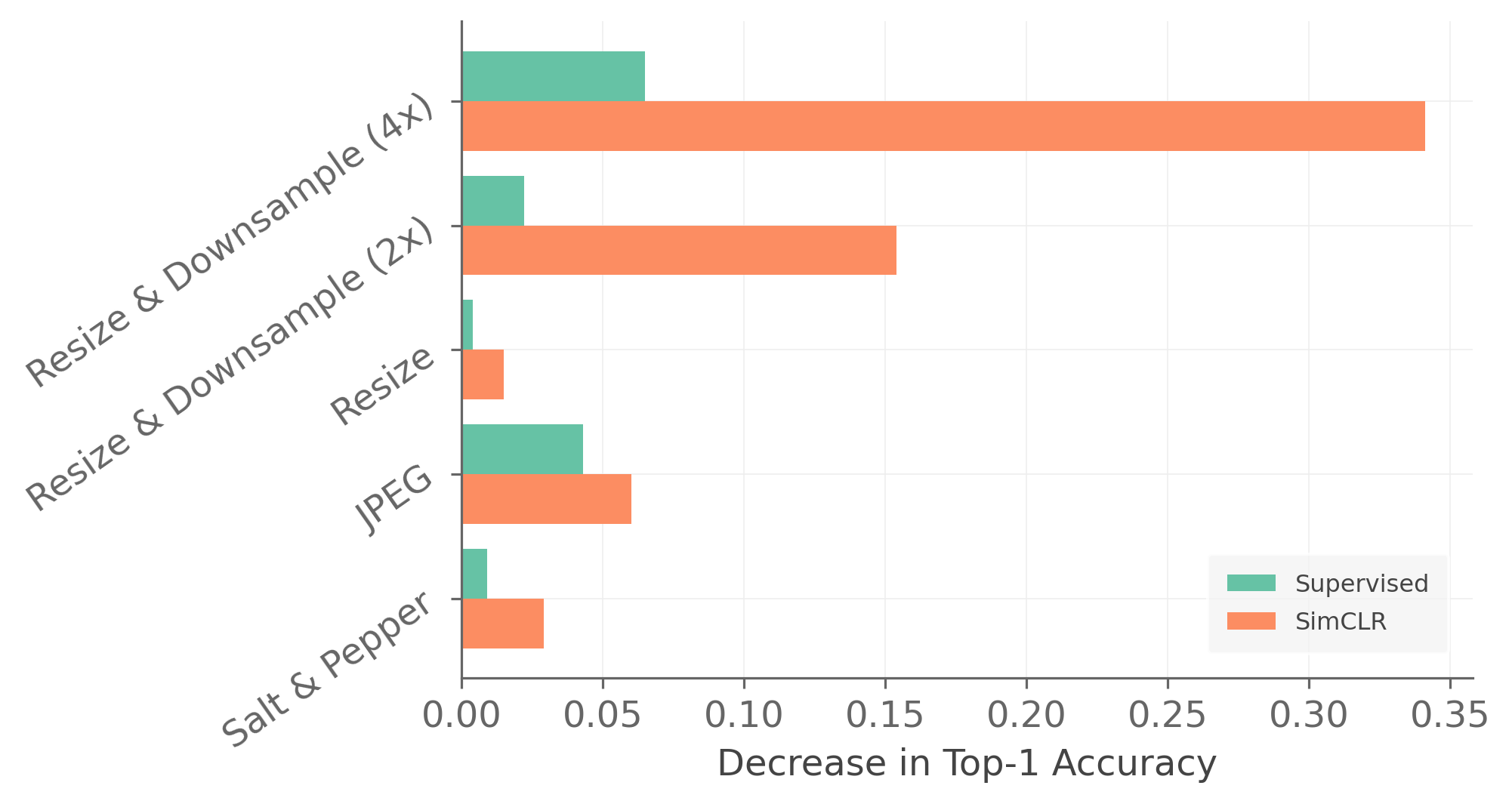}
    \vspace{-8pt}
    \caption{
    \textbf{What is the effect of pretraining image corruption?}
    Decrease in linear evaluation accuracy on ImageNet due to pretraining on corrupted versions of the ImageNet training set. 
    The zero point corresponds to pretraining (supervised or SimCLR) on uncorrupted images followed by linear evaluation.
    ``Supervised" and ``SimCLR" have different zero points. 
    All linear classifiers are trained using the full uncorrupted ImageNet training set.
    }
    \vspace{-12pt}
    \label{fig:image_corruption}
\end{figure}

In Figure~\ref{fig:image_corruption} we present linear evaluation results on ImageNet for a simple but diverse set of corruptions. 
The zero point corresponds to pretraining on uncorrupted images, and we measure how much performance drops when pretraining on corrupted images. 
The ``Salt and Pepper" corruption is salt and pepper noise applied independently to each pixel, in each channel, with probability 0.01.
The ``JPEG" corruption is JPEG compression with a very low quality level of 10.
For ``Resize", we resize each image so that the short side is 256 pixels while preserving the aspect ratio. 
This reduces the resolution of the crops used for training. 
For our downsampling corruptions, we follow the resize operation with downsampling by 2x or 4x and then upsampling by the same factor. 
This holds constant the image size and the fraction of the image occupied by each object, but reduces resolution. 
Implementation details and examples can be found in the supplementary.

\noindent
\textbf{Image resolution is critical for SSL.} 
``Downsample (2x)" and ``Downsample (4x)" are by far the most damaging corruptions for SimCLR, reducing accuracy by around $15\%$ and $34\%$, respectively. 
Since SimCLR already involves extreme cropping, we might expect more robustness to changes in image resolution. 
This finding could be partially explained by the difficulty of generalizing to higher-resolution images during linear classifier training \cite{touvron2019fixing}. 
However, supervised pretraining faces the same challenge but the effect of downsampling is much less dramatic.
This suggests that the performance drop is due to deficiencies in the features learned by SimCLR. 

\noindent
\textbf{SSL is relatively robust to high-frequency noise.}
``JPEG" and ``Salt \& Pepper" both add high-frequency noise to the image. 
For SimCLR, these corruptions have a much milder impact than the downsampling corruptions.
One possible explanation is that downsampling destroys texture information, which is known to be a particularly important signal for convolutional neural networks~\cite{geirhos2018imagenet, hermann2019origins}. 
For supervised pretraining the ranking of corruptions is very different, with ``JPEG"  landing between 2x and 4x downsampling. 

\subsection{Task granularity}\label{sec:task_granularity}

\begin{figure*}
    \centering
    \includegraphics[width=0.33\textwidth]{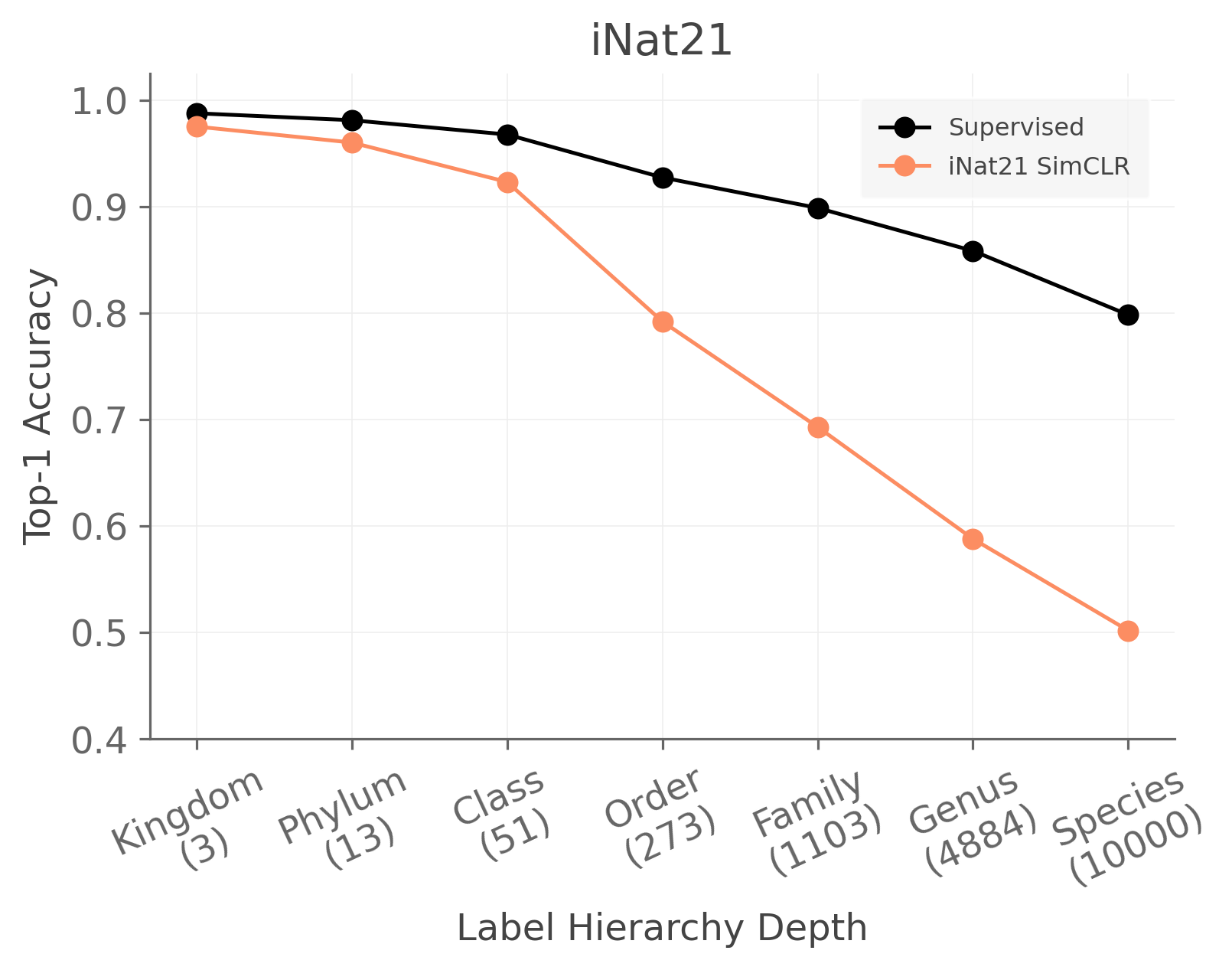}
    \includegraphics[width=0.33\textwidth]{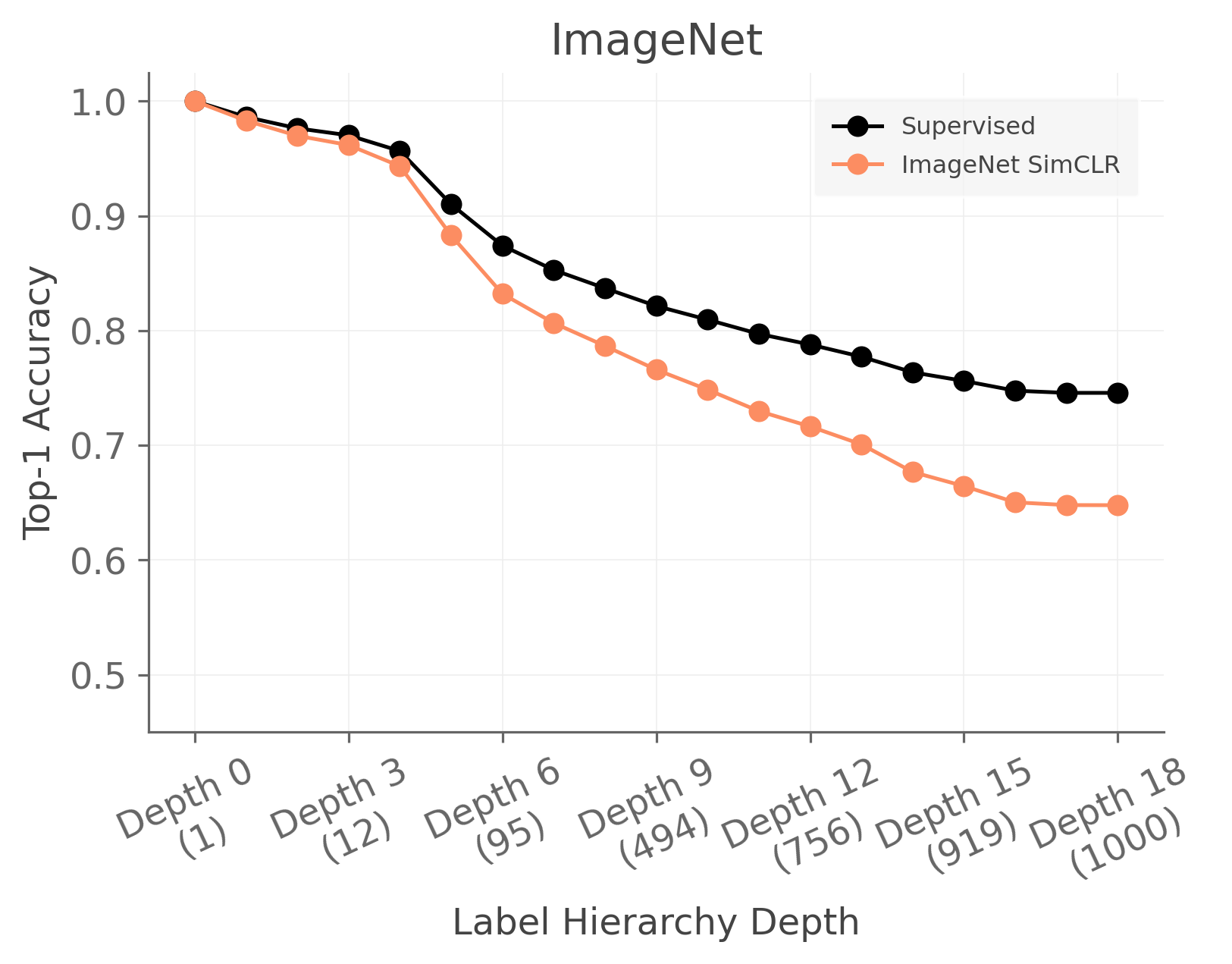}
    \includegraphics[width=0.33\textwidth]{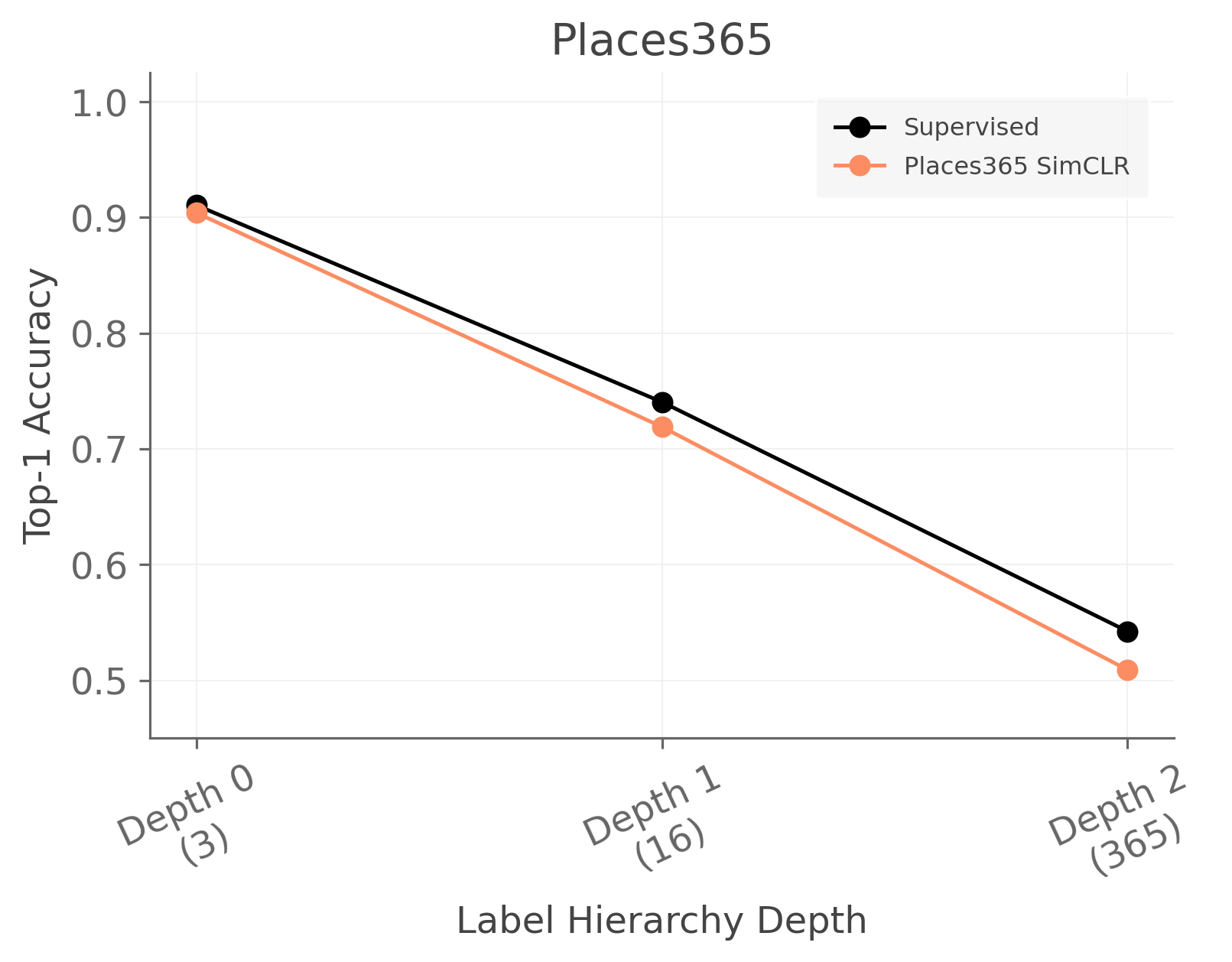}
    \vspace{-20pt}
    \caption{
    \textbf{How does performance depend on label granularity?}
    Linear evaluation at different levels of label granularity for iNat21, ImageNet, and Places365.
    Each plot compares supervised learning from scratch against a linear classifier trained on top of in-domain SimCLR.
    Both are trained using the full training sets.
    We plot top-1 accuracy against label granularity, which is more fine-grained as we move from left to right.
    The numbers on the $x$-axis are the class counts at a given level of the label hierarchy.
    We do not re-train at coarser granularity levels, we just change the evaluation label set.
    The definitions of the hierarchy levels are given in the supplementary material.
    }
    \label{fig:performance_granularity}
    \vspace{-12pt}
\end{figure*}

We have seen that the properties of pretraining datasets are important for determining the utility of self-supervised representations. 
But are there downstream tasks for which self-supervised representations are particularly well or poorly suited? 
We consider \emph{fine-grained classification} and show that classification performance depends on \emph{task granularity}, \ie how fine or coarse the labels are. While there are formal methods for measuring dataset granularity \cite{cui2019measuring}, we claim by intuition that iNat21 is more fine-grained than ImageNet, which is more fine-grained than Places365. 

In Figure~\ref{fig:performance_granularity} we use label hierarchies (which are available for ImageNet, iNat21, and Places365) to explicitly study how performance depends on label granularity.
We treat ``distance from the root of the hierarchy" as a proxy for granularity, so labels further from the root are considered to be more fine-grained. 
We perform (i) linear classifier training (for SimCLR) and (ii) end-to-end training from scratch (for ``Supervised") using the labels at the finest level of the taxonomy and re-compute accuracy values as we progressively coarsen the predictions and labels. 
We do not re-train at each level of granularity.
A complete description of this process can be found in the supplementary materials. 

\noindent
\textbf{The performance gap between SSL and supervised learning grows as task granularity becomes finer.} 
We start with the iNat21 results in Figure~\ref{fig:performance_granularity}.
The supervised and SimCLR pretrained models perform similarly at the coarsest levels of the label hierarchy (``Kingdom"). 
Both models perform worse as task granularity increases, but the SimCLR model degrades much more rapidly (``Species"). 
This suggests that SimCLR may fail to capture fine-grained semantic information as effectively as supervised pretraining.
We also observe a growing supervised/self-supervised gap for ImageNet and Places365. 
The magnitude of this gap seems to track dataset granularity, since iNat21 (most fine-grained) has the largest gap and Places365 (least fine-grained) has the smallest gap. 
The fact that supervised learning achieves high performance on iNat21 while SSL lags behind suggests that iNat21 could be a valuable benchmark dataset for the next phase of SSL research. 

\noindent
\textbf{Are the augmentations destructive?} 
State-of-the-art contrastive learning techniques are designed for ImageNet, so the default augmentation policy may be poorly tuned for other datasets~\cite{xiao2020should}. 
For instance, if color is a key fine-grained feature for species classification then the ``color jitter" augmentation used by SimCLR may destroy important information for iNat21 classification.
Could this explain the rapid drop in performance exhibited by iNat21 SimCLR for fine-grained classes? 
Notice that there is a similar, though less extreme, fine-grained performance drop for ImageNet SimCLR in Figure~\ref{fig:performance_granularity}.
Since the ImageNet-tuned augmentations are presumably not destructive for ImageNet, it does not seem likely that this fully explain our observations.

\noindent
\textbf{Does contrastive learning have a coarse-grained bias?} 
We hypothesize that the contrastive loss tends to cluster images based on overall visual similarity. 
The intuition is that fine-grained features are often subtle, and subtle features are unlikely to be very useful for distinguishing between pairs of images in the contrastive pretext task. 
If our hypothesis is correct then the boundaries between different clusters would not be well-aligned with the boundaries between fine-grained classes. 
This effect could be overlooked when evaluating on coarse-grained classes, but would become apparent on a more fine-grained task. 
Additional analysis is required to fully understand this ``granularity gap" in SSL, which we leave to future work. 

\section{Conclusion}

We have presented a comprehensive set of experiments to address several aspects of the question: \emph{when does contrastive visual representation learning work?} In Section~\ref{sec:data_quantity} we found that we need fewer than 500k pretraining images before encountering severe diminishing returns. 
However, even the best self-supervised representations are still much worse than peak supervised performance without hundreds of thousands of labeled images for classifier training.
In Section~\ref{sec:data_domain} we found that self-supervised pretraining on 1M images from different domains results in representations with very different capabilities, and that simple methods for combining different datasets do not lead to large gains. 
In Section~\ref{sec:data_quality} we showed that image resolution is critical for contrastive learning and, more broadly, that some image corruptions can degrade a self-supervised representation to the point of unusability while others have almost no impact. 
Finally, in Section~\ref{sec:task_granularity} we found that supervised pretraining retains a substantial edge when it comes to fine-grained classification.
These experiments highlight several areas where further research is needed to improve current SSL algorithms, most of which were not evident from traditional evaluation protocols, \ie top-1 accuracy on ImageNet.

\noindent
\textbf{Limitations.} 
We mainly perform experiments using one self-supervised method. 
We focus on SimCLR because it reflects the essence of state-of-the-art contrastive learning methods without introducing additional architectural complexities.
While our MoCo and BYOL experiments are not much different from SimCLR, it is important to validate our results on other self-supervised methods. 
It would also be interesting to explore alternative backbone architectures~\cite{dosovitskiy2020image,caron2021emerging}, though after controlling for training settings, ResNet-50 remains competitive with newer architectures~\cite{xiao2021early,wightman2021resnet}. 
We study only classification tasks, so additional work is also required to understand how these results translate to segmentation~\cite{wang2021exploring} or detection~\cite{Zoph2020-fi,henaff2021efficient}.
Finally, we only consider datasets up to roughly ImageNet scale. We believe this is the most practical setting for most use cases, but it is possible that some patterns may be different for significantly larger datasets and models \cite{goyal2021self, goyal2022vision}.

\noindent
\textbf{Acknowledgements.}
We thank Mason McGill for detailed feedback, and Grant Van Horn, Christine Kaeser-Chen, Yin Cui, Sergey Ioffe, Pietro Perona, and the rest of the Perona Lab for insightful discussions. 
This work was supported by the Caltech Resnick Sustainability Institute, an NSF Graduate Research Fellowship (grant number DGE1745301), and the Pioneer Centre for AI (DNRF grant number P1).

{\small
\bibliographystyle{ieee_fullname}
\bibliography{main}
}

\clearpage
\appendix
\setcounter{table}{0}
\renewcommand{\thetable}{A\arabic{table}}
\setcounter{figure}{0}
\renewcommand{\thefigure}{A\arabic{figure}}
\section{Additional Results}\label{sec:additional_results}

\subsection{How does task granularity affect different self-supervised learning methods?}
\begin{figure}
    \centering
    \includegraphics[width=0.475\textwidth]{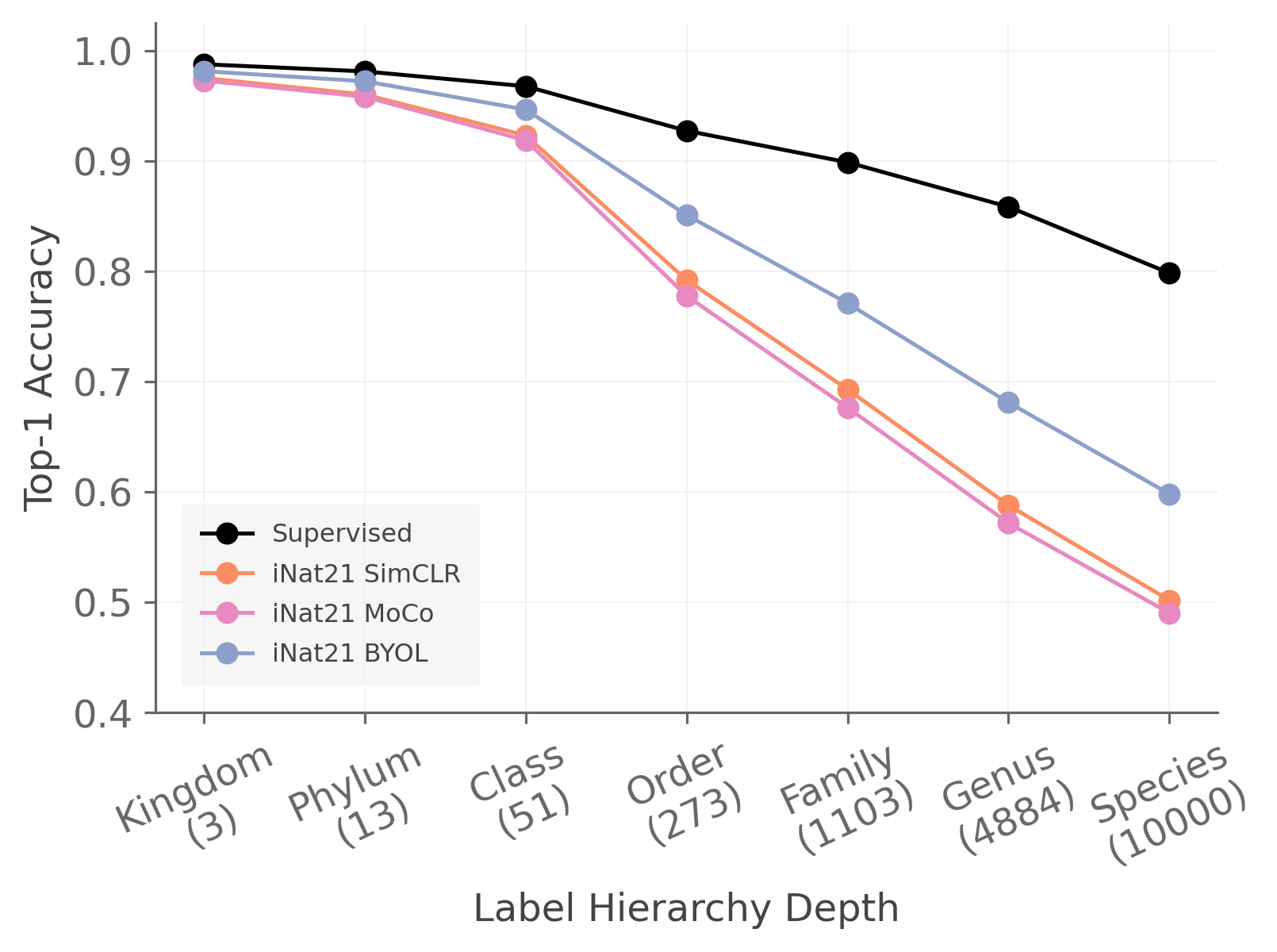}
    \caption{
    \textbf{How does performance depend on label granularity?} Linear evaluation at different levels of label granularity for iNat21. We compare end-to-end training from scratch against linear classifiers trained on top of in-domain self-supervised representations (SimCLR, MoCo, and BYOL). 
    All classifiers (linear and end-to-end) are trained using the full iNat21 training set. 
    This plot is identical to Figure~\ref{fig:performance_granularity} except that we have added curves for MoCo and BYOL. 
    }
    \label{fig:task_granularity_ssl_comparison}
\end{figure}

In Figure~\ref{fig:performance_granularity} we saw that there is a large gap between supervised and self-supervised (SimCLR) performance on iNat21. 
Figure~\ref{fig:task_granularity_ssl_comparison} extends Figure~\ref{fig:performance_granularity} by adding results for MoCo and BYOL. 
Across all granularity levels, MoCo is slightly worse than SimCLR and BYOL is significantly better. 
For all three self-supervised methods, performance drops rapidly as the evaluation is made more fine-grained. 
While BYOL is much better than SimCLR, it still lags behind fully supervised performance by 20\% top-1 accuracy.

\subsection{Do larger models scale better in terms of pretraining set size?}
\begin{figure}
    \centering
    \includegraphics[width=0.475\textwidth]{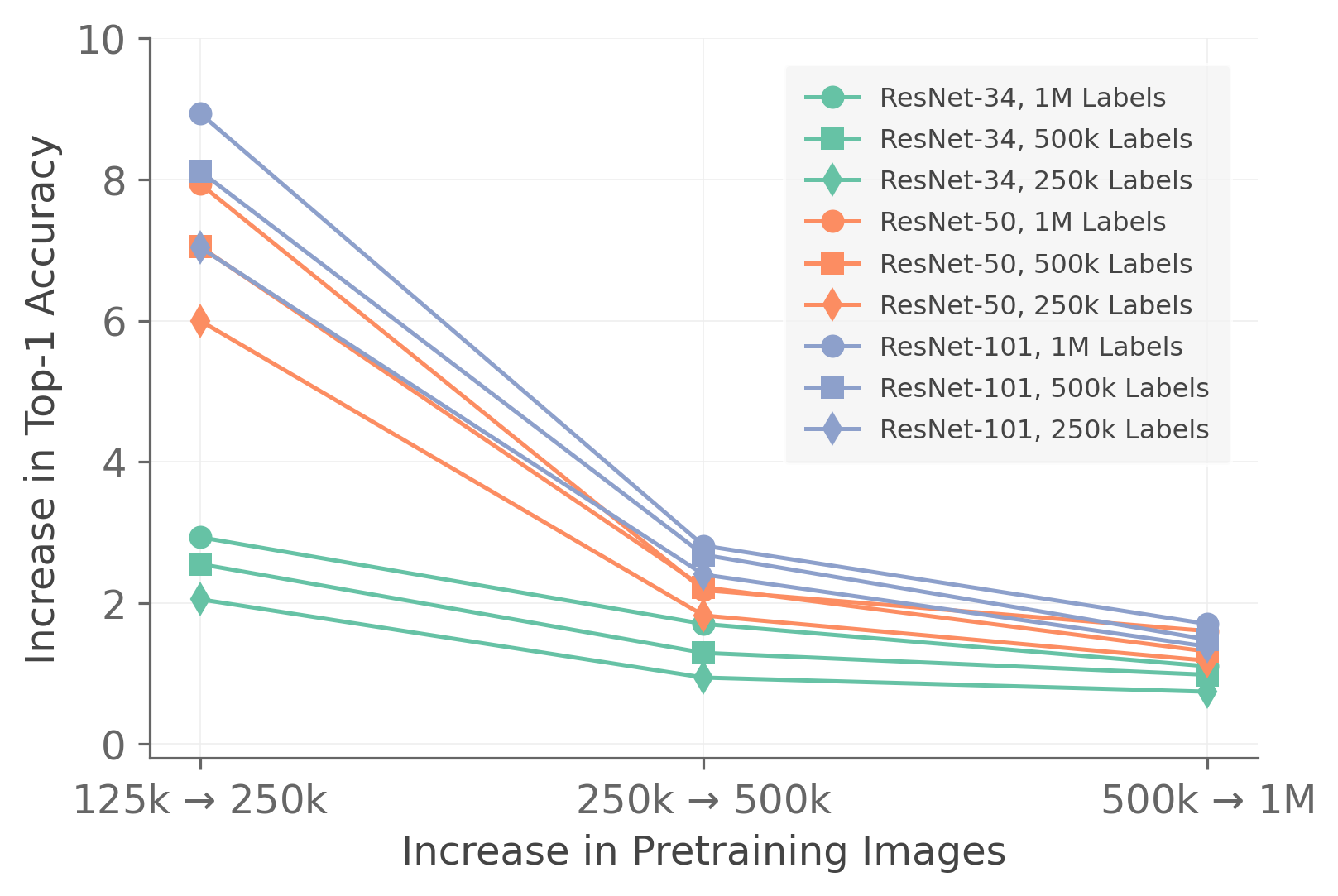}
    \caption{
    \textbf{Increasing pretraining set size leads to rapid diminishing returns across different model sizes.} 
    Linear evaluation results on iNat21 for SimCLR. 
    We show the increase in top-1 accuracy on iNat21 that results from doubling pretraining set size. 
    Each color is a different architecture. 
    For a given color, each line uses a different amount of labeled data for linear classifier training. 
    }
    \label{fig:model_sizes}
\end{figure}

In Figure~\ref{fig:performance_vs_num_images} we observe that doubling the pretraining set size from 500k images to 1M images leads to small benefits (1-2\%) across three large-scale datasets. 
However, all of those results are based on a ResNet-50. 
Does the story change for larger or smaller models? 
In Figure~\ref{fig:model_sizes} we study this question using ResNet-34, ResNet-50, and ResNet-101.
When we double the size of the pretraining set from 125k to 250k, ResNet-50 and ResNet-101 make significantly larger gains than ResNet-34. 
However, doubling the size of the pretraining set from 500k to 1M produces gains of $<$2\% for all models. 
While ResNet-101 gains more than ResNet-50 with each increase in pretraining set size, the gap between them is very small by the time we reach 1M images. 
This is the same conclusion we reached in Figure~\ref{fig:performance_vs_num_images}.

\subsection{Does semantic similarity explain patterns in self-supervised performance?}

In Section~\ref{sec:data_domain} we saw that (i) in-domain SimCLR pretraining always beats cross-domain SimCLR pretraining and (ii) ImageNet is the best dataset for cross-domain pretraining. 
One hypothesis which could explain these patterns is that \emph{semantic similarity between the pretraining dataset and the downstream task leads to better performance.}
This would require that modern self-supervised methods capture high-level semantic information. 
In this section we consider evidence for this hypothesis.

\begin{figure}
    \centering
    \includegraphics[width=0.475\textwidth]{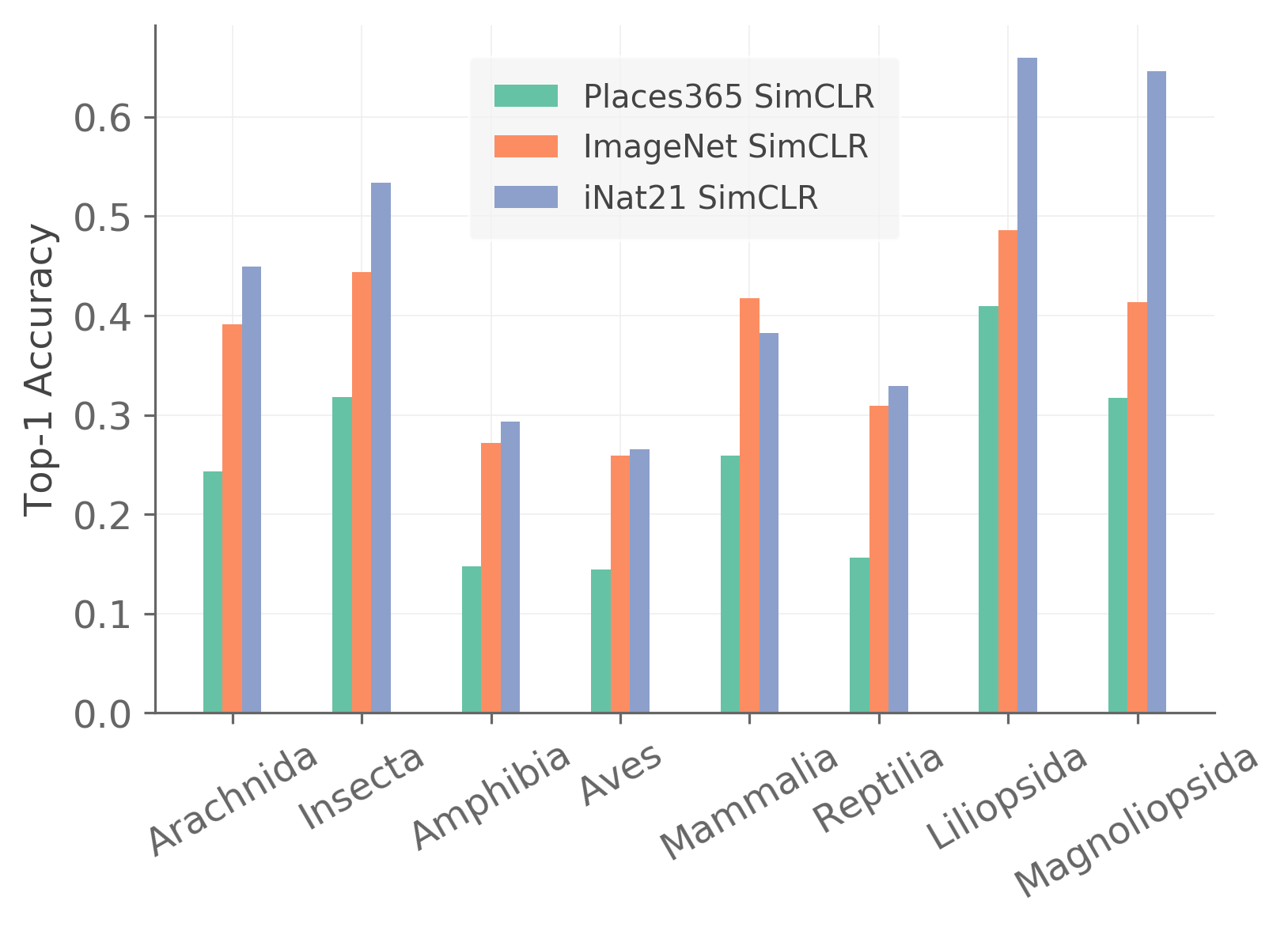}
    \caption{
    \textbf{Semantic similarity may predict transfer performance.}
    Linear evaluation results on iNat21 for different pretrained representations. 
    We compare representations pretrained on Places365, ImageNet, and iNat21 (full datasets, not subsampled) in terms of top-1 linear classification accuracy on iNat21.
    The result for each taxonomic class (Arachnida, Insecta, Amphibia, Aves, Mammalia, Reptilia, Liliopsida, Magnoliopsida) is the average of the per-species accuracy over all species in that taxonomic class. 
    }
    \label{fig:inat21_class_increase}
\end{figure}

\noindent
\textbf{ImageNet SimCLR performs well on iNat21 classes that are similar to ImageNet classes.}
ImageNet includes around 200 mammal categories, 60 bird categories, and 30 categories of insects and reptiles. 
A breakdown of the categories in iNat21 can be found in \cite{van2021benchmarking}.
In Figure~\ref{fig:inat21_class_increase} we analyze per-categories accuracy averaged over six \emph{taxonomic classes} of animals (Arachnida, Insecta, Amphibia, Mammalia, Reptilia) and two taxonomic classes of plants (Liliopsida and Magnoliopsida).
Surprisingly, ImageNet SimCLR outperforms iNat21 SimCLR on mammals (Mammalia) and nearly matches the performance of iNat21 SimCLR on birds (Aves).
We also evaluate Places365 SimCLR pretraining, which does not have any categories corresponding to animals or plants. 
We do not see any taxonomic classes for which Places365 SimCLR performs close to iNat21 SimCLR.

\noindent
\textbf{Most of the ImageNet classes for which iNat21 SimCLR beats ImageNet SimCLR are animals or plants.} 
We find similar effects in the context of ImageNet classification. 
When we compare the per-category accuracy for ImageNet SimCLR with the per-category accuracy for iNat21 SimCLR, we find that ImageNet SimCLR leads to a higher accuracy for all but 80 categories. Of those 80 categories, 68 (i.e. 85\%) are animals or plants. 

\begin{figure}
    \centering
    \includegraphics[width=0.45\textwidth]{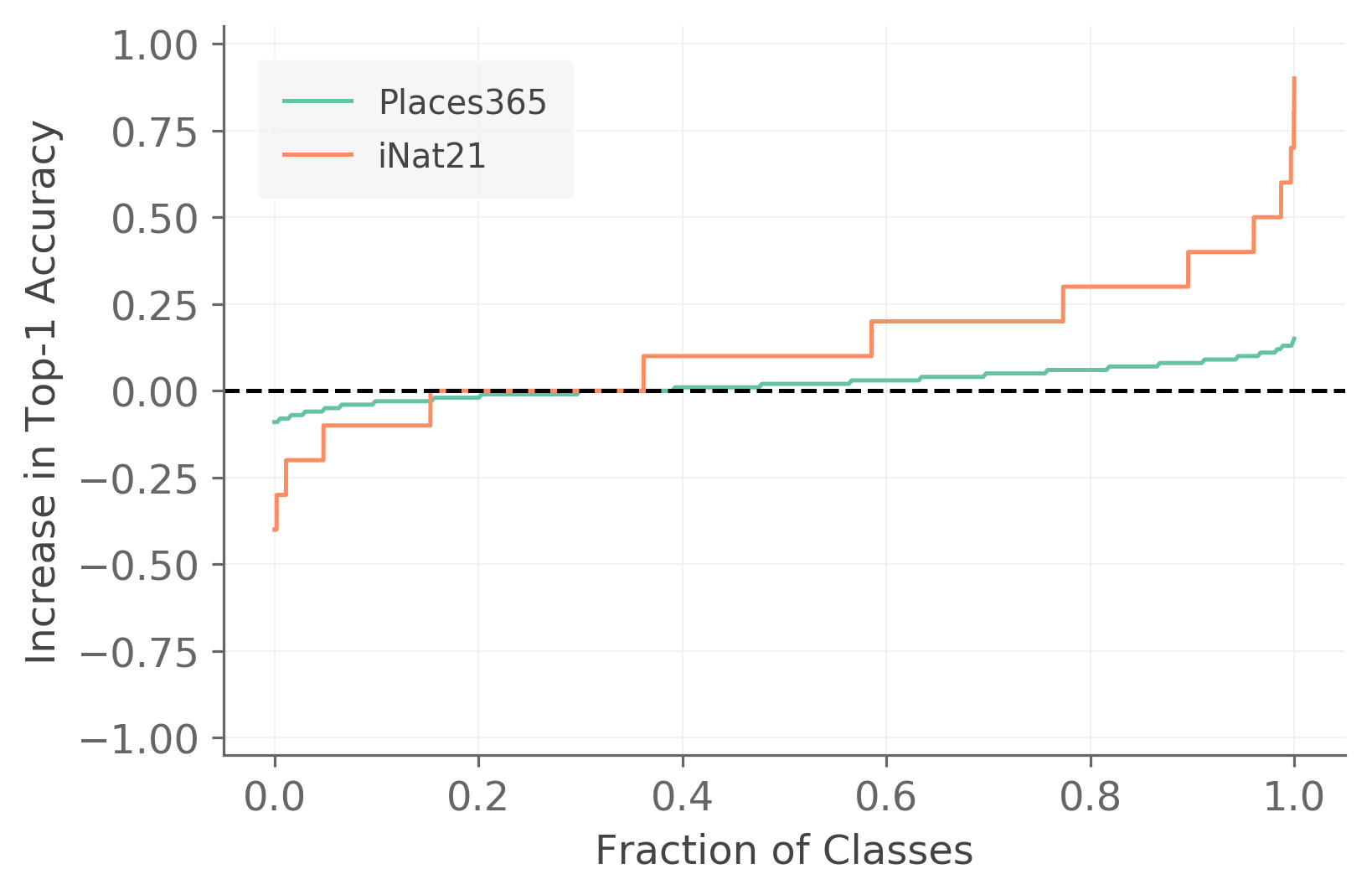}
    \caption{
    \textbf{In-domain contrastive learning improves accuracy on most (but not all) classes.}
    Increase in per-class linear evaluation results for different pretrained representations compared to an ImageNet SimCLR baseline.
    In Table~\ref{tab:large_dataset_grid_simclr} we saw that in-domain pretraining was better than cross-domain pretraining. 
    Here we break down those results in terms of the per-class accuracy increase for in-domain SimCLR with respect to ImageNet SimCLR (represented by the dashed line).
    For both Places365 (green line) and iNat21 (orange line), in-domain SimCLR pretraining benefits around 60\% of classes while around 40\% of classes are either the same or worse off.
    Note that the curves for Places365 and iNat21 are sorted independently so the species ordering is different for each.
    See Table~\ref{tab:improved_harmed} for lists of the most harmed and most improved classes for both datasets. 
    }
    \label{fig:class_acc_difference}
\end{figure}

\noindent 
\textbf{In-domain pretraining helps some classes and hurts others.}
To develop a deeper understanding of the effect of pretraining domain, we compute the per-class accuracy improvement that results from using in-domain SimCLR instead of ImageNet SimCLR. 
We present these results for iNat21 and Places365 in Figure~\ref{fig:class_acc_difference}. 
For these results we pretrain on the full datasets, not the million-image subsets. 
We see that in-domain pretraining leads to an improvement for $\sim 60\%$ of classes, while the rest stay the same or degrade.
In Table~\ref{tab:improved_harmed}  we list the most harmed and most improved classes. 
Interestingly, all of the most improved classes for iNat21 are plants. Around 40\% of the images in iNat21 are plants, but of course the self-supervised method does not have access to the labels.
We also notice that many of the most harmed classes for iNat21 are similar to classes we might find in ImageNet, e.g. birds, mammals, and reptiles. 
This is consistent with the hypothesis that the success of SimCLR is partially governed by the semantic similarity between the pretraining set and the downstream task, even though no labels are used for representations learning.
The patterns seem less clear for Places365.

\begin{table*}[t]
\centering
\scriptsize
\begin{tabular}{c c|c c}
\multicolumn{2}{c}{iNat21} & \multicolumn{2}{c}{Places365} \\
Most Improved & Most Harmed & Most Improved & Most Harmed \\ \hline
\verb|summer-cypress| & \verb|Ferruginous Hawk| & \verb|/a/airport_terminal| & \verb|/s/slum| \\
\verb|Greater Tickseed| & \verb|Western Banded Gecko| & \verb|/r/roof_garden| & \verb|/h/home_office| \\
\verb|Annual Blue-eyed Grass| & \verb|Desert Cottontail| & \verb|/r/restaurant| & \verb|/s/swamp| \\
\verb|Jamaica Snakeweed| & \verb|Arizona Alligator Lizard| & \verb|/g/gazebo/exterior| & \verb|/a/arena/performance| \\
\verb|California Jacob's ladder| & \verb|Petticoat Mottlegill| & \verb|/b/bedroom| & \verb|/b/beach| \\
\verb|tomato| & \verb|Elk| & \verb|/b/booth/indoor| & \verb|/c/canal/urban| \\
\verb|leatherleaf fern| & \verb|Ruddy Ground-Dove| & \verb|/r/rice_paddy| & \verb|/o/orchard| \\
\verb|northern bugleweed| & \verb|Long-tailed Weasel| & \verb|/c/castle| & \verb|/o/ocean| \\
\verb|mock azalea| & \verb|Little Blue Dragonlet| & \verb|/m/museum/indoor| & \verb|/g/garage/indoor| \\
\verb|Mexican False Calico| & \verb|Signal Crayfish| & \verb|/l/locker_room| & \verb|/u/underwater/ocean_deep| 
\end{tabular}
\vspace{2pt}
\caption{
\textbf{In-domain pretraining helps some classes and harms others.}
Lists of the ten most improved and the ten most harmed classes when we change from ImageNet SimCLR pretraining to in-domain SimCLR pretraining. 
See Figure~\ref{fig:class_acc_difference} for the corresponding curves showing the distribution of accuracy improvement over all classes.
}
\label{tab:improved_harmed}
\end{table*}

\subsection{Is SimCLR overly tuned for ImageNet?}\label{sec:simclr_overfit_imagenet}
One possible explanation for the strong cross-domain performance of ImageNet SimCLR we observe in Table~\ref{tab:large_dataset_grid_simclr} is that the training procedures, augmentations, and hyperparameters for SimCLR were designed with ImageNet in mind. 
This might lead SimCLR to produce better representations when trained on ImageNet than it does when trained on other datasets. 
However, we see that in-domain SimCLR is better than ImageNet SimCLR for iNat21, Places365, and GLC20.
If SimCLR is somehow ``overfit" to ImageNet, that effect seems to be overwhelmed by the effect of domain similarity.

\subsection{What is the effect of native image resolution?}\label{sec:native_image_resolution}
ImageNet and iNat21 have larger images than Places365 and GLC20. 
While images are always resized to 224x224 before they are passed in to the network, that happens after random crops are chosen. 
This means that we are training on more detailed 224x224 images for ImageNet and iNat21 compared to Places365 and GLC20.
This could affect cross-domain performance comparisons such as those in \eg Table~\ref{tab:large_dataset_grid_simclr}. 
To understand the impact of this difference, we compare pretraining on resized images to pretraining on the original images for ImageNet and iNat21. 
We provide linear evaluation results in Table~\ref{tab:image_size}. 
It seems that resizing can introduce a 1-2\% difference in top-1 accuracy, which can be significant on datasets like ImageNet where the performance improvements of new methods are also on the order of 1-2\%. 

\begin{table}[tb]
\centering
\scriptsize
\begin{tabular}{l|c|c|c|c}
Pretraining & iNat21 & ImageNet & Places365 & GLC20\\ \hline
iNat21 & 0.506 & 0.520 & 0.413 & 0.865\\
iNat21 (Resize) & 0.505 & 0.500 & 0.412 & 0.865 \\
Change & -0.001 & -0.020 & -0.001 & 0.000 \\ \hline
ImageNet & 0.380 & 0.647 & 0.488 & 0.710\\ 
ImageNet (Resize) & 0.394 & 0.632 & 0.471 & 0.712 \\
Change & +0.014 & -0.015 & -0.017 & +0.002
\end{tabular}
\vspace{-6pt}
\caption{
\textbf{Analysis of the effect of native image size.}
Linear evaluation results for representations pretrained on resized versions of ImageNet and iNat21. 
ImageNet and iNat21 have images that vary in size, many of which are much larger than the 256x256 images in Places365 and GLC20. 
Here we analyze the effect of pretraining on resized variants of ImageNet and iNat, which have been preprocessed so that all images have a short side of 256. 
We use the ``Resize" corruption described in Appendix~\ref{sec:implementation_details}. 
Note that the downsampling results in Figure~\ref{fig:image_corruption} start from resized datasets -- in this table we are analyzing the effect of the initial resizing.
} 
\label{tab:image_size}
\end{table}

\subsection{Is class difficulty preserved between different representations?}\label{sec:class_difficulty}
\begin{figure*}
    \begin{subfigure}{\textwidth}
        \centering
        \captionsetup{justification=centering}
        \includegraphics[width=0.45\textwidth]{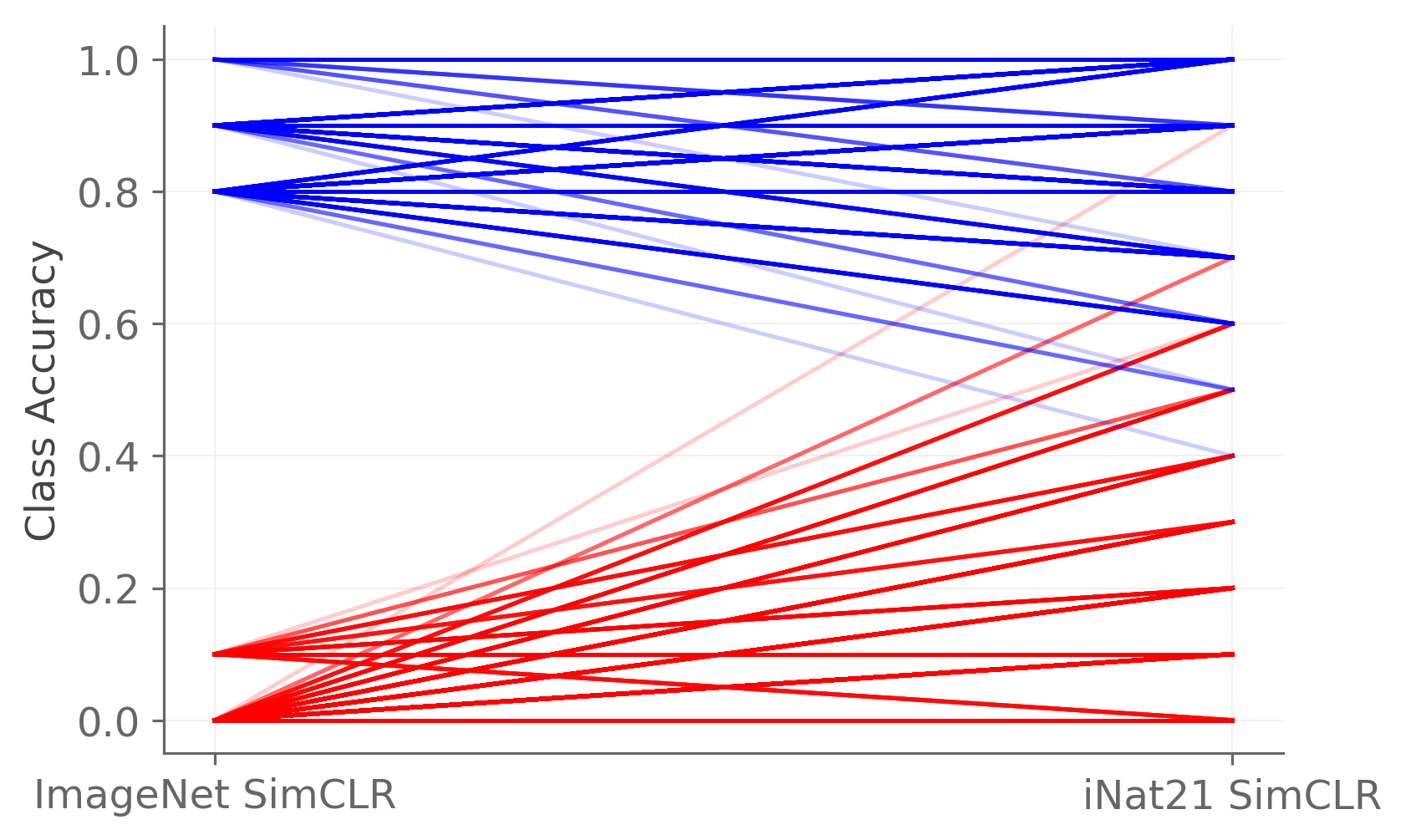}
        \includegraphics[width=0.45\textwidth]{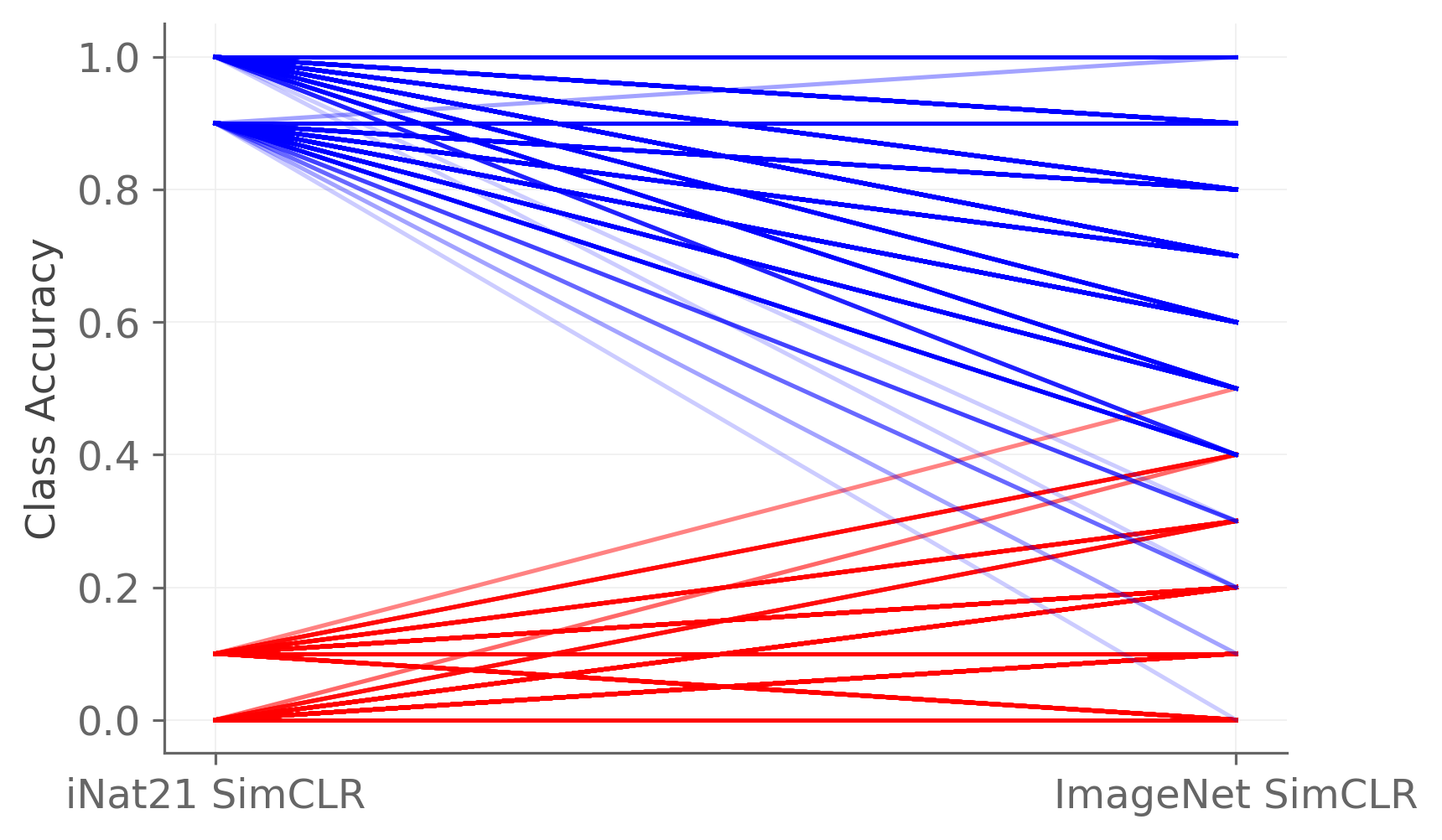}
        \caption{Results for iNat21 classification.}
    \end{subfigure}
    \begin{subfigure}{\textwidth}
        \centering
        \captionsetup{justification=centering}
        \includegraphics[width=0.45\textwidth]{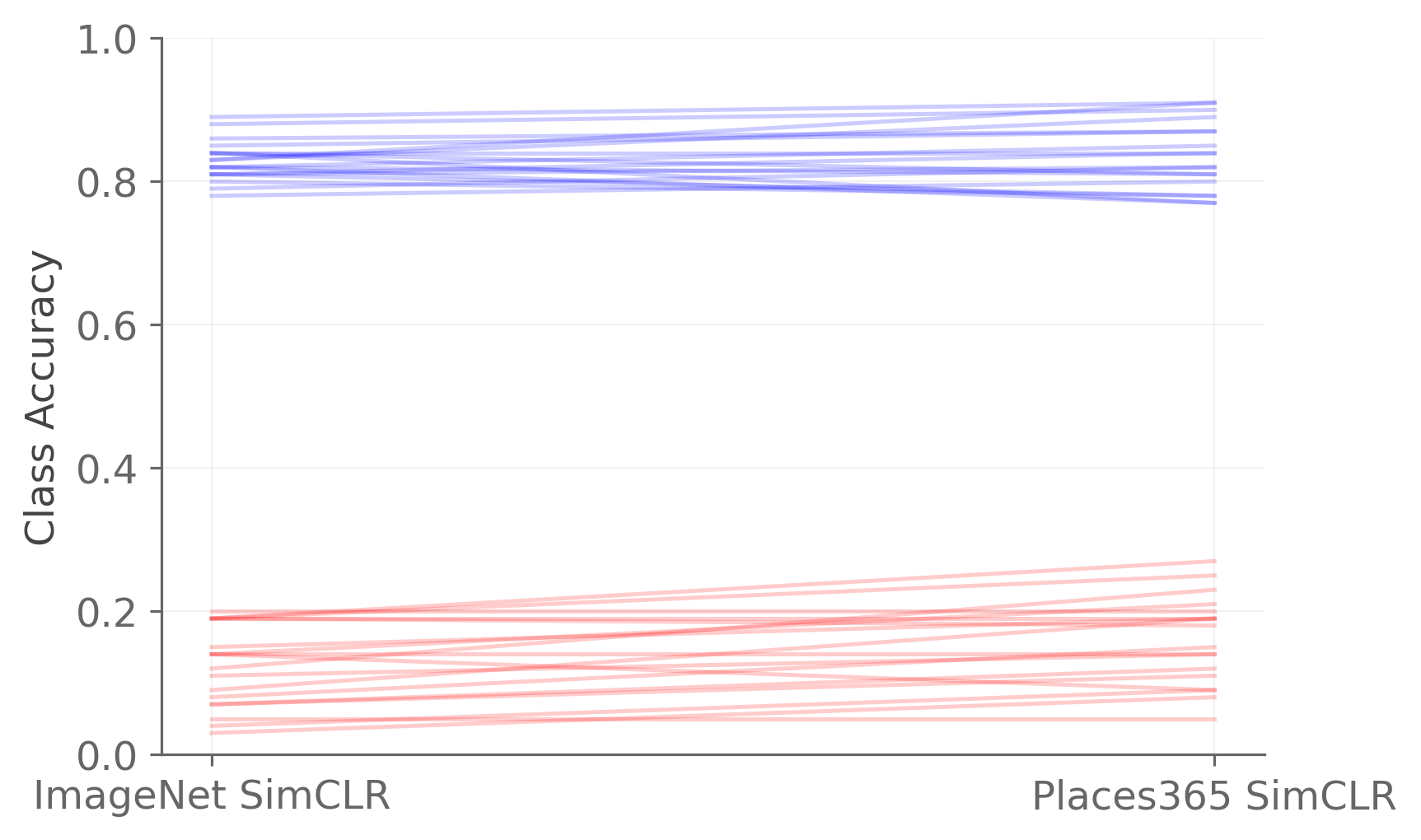}
        \includegraphics[width=0.45\textwidth]{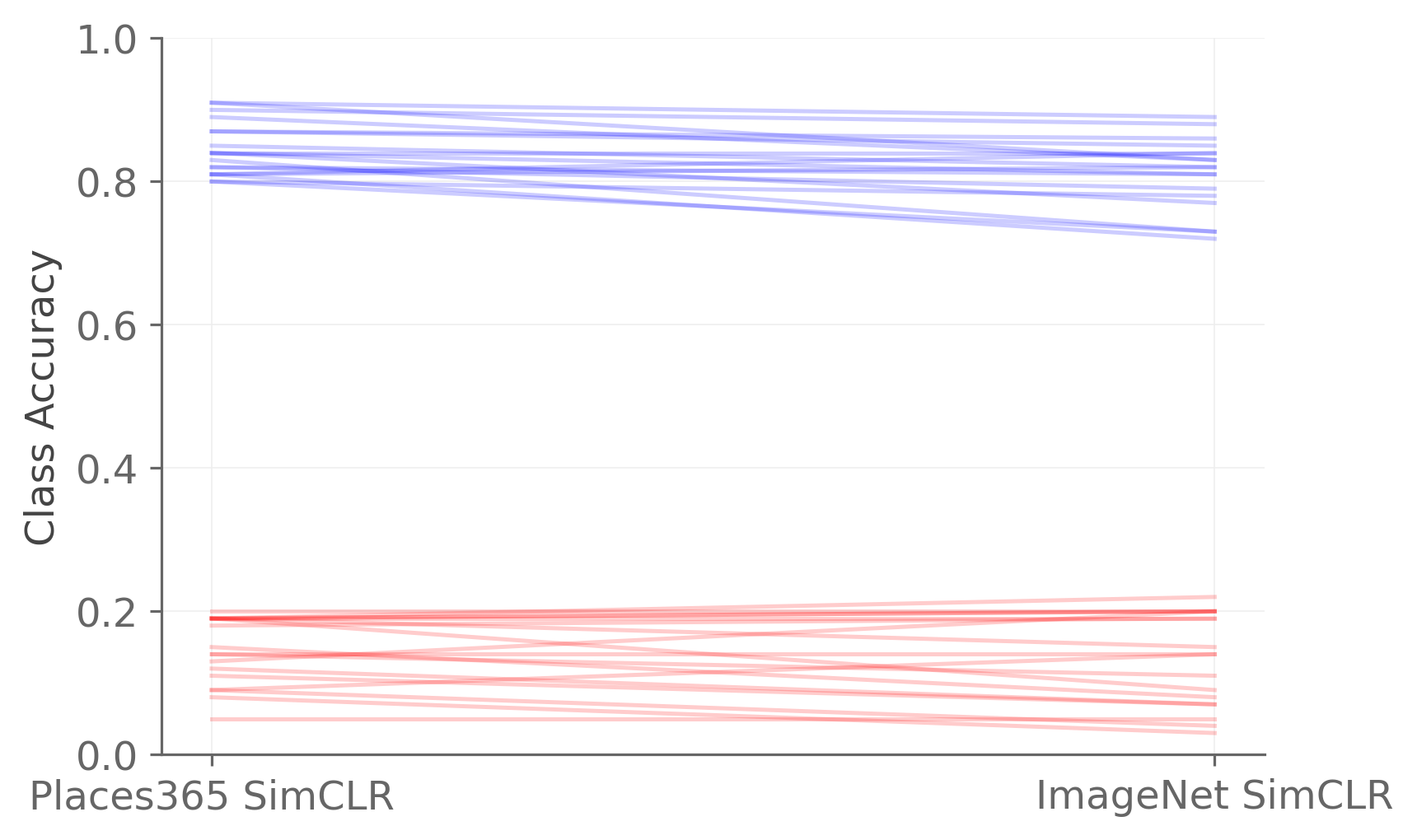}
        \caption{Results for Places365 classification.}
    \end{subfigure}
    \caption{
    \textbf{Difficulty depends on the representation.} 
    Visualization of the change in per-class linear evaluation results when the underlying self-supervised representation is changed.
    We show the hardest 5\% of classes (red lines) and the easiest 5\% of classes (blue lines) for the representation named in the bottom left corner of each panel.
    Left and right plots simply reverse which representation is being used to define the easy and hard classes. 
    Each line represents one class, and shows how the accuracy for that class increases or decreases when we replace the representation named in the bottom-left corner of each panel with the representation named in the bottom-right corner of each panel. 
    Note that the iNat21 validation set has 10 images per class, so all class accuracy values for the top plots lie in $\{0, 0.1, \ldots, 0.9, 1.0\}$.
    }
    \label{fig:class_difficulty comparison}
\end{figure*}

To analyze the differences between self-supervised representations a bit further, we ask whether the same classes are ``difficult" or ``easy" under different representations.
In Figure~\ref{fig:class_difficulty comparison} we illustrate how per-class accuracy changes for iNat21 (top row) and Places365 (bottom row) when switching between ImageNet SimCLR and in-domain SimCLR.
The panels in the left column define the hardest and easiest examples based on ImageNet SimCLR, while the panels in the right column define the hardest and easiest examples based on in-domain SimCLR. 
We observe that class difficulty is not preserved between ImageNet SimCLR and iNat21 SimCLR (top row), but it is largely preserved between ImageNet SimCLR and Places365 SimCLR (bottom row). 
We also note that the overall patterns are the same whether we track the easiest/hardest examples for ImageNet SimCLR and move to the in-domain representation (left column) or track the easiest/hardest examples for in-domain SimCLR and move to ImageNet SimCLR (right column). 

\subsection{What is the effect of within-dataset diversity?}\label{sec:intra_dataset_diversity}
In Table~\ref{tab:pooled_data} we saw that adding pretraining images from a different dataset provides little to no benefit whereas adding pretraining images from the same dataset consistently helps. 
The surprising conclusion is that a larger, more diverse pretraining dataset can be worse than a smaller, homogeneous pretraining dataset.  
In this section we present a preliminary study of a milder form of data diversity by changing the number of classes in our pretraining data while holding the number of images constant. 
We construct three equally sized subsets of ImageNet: one with 200 classes (500 images per class), one with 500 classes (200 images per class), and one with 1k classes (100 images / class).
We present linear evaluation results in Table~\ref{tab:data_diversity}.
The class information is only used to construct the datasets, which are then used for self-supervised pretraining.
Linear classifiers trained on top of these representations use full training sets as usual.

If we assume that class count is a valid proxy for visual diversity, then Table~\ref{tab:data_diversity} indicates that increasing diversity improves performance on Places365 (+3.9\% top-1) but degrades performance on iNat21 (-2.4\% top-1). All else being equal, we might intuitively expect a more diverse pretraining set to be beneficial. This seems to be the case for Places365. However, the result for iNat21 shows that this is not necessarily the case. It is possible that more homogeneous pretraining data leads to more fine-grained self-supervised features, which would account for the decrease in performance with increasing diversity for iNat21. Since Places365 is not very fine-grained, it would not benefit from this effect. 
However, this is a small-scale experiment on one dataset so it should be interpreted with caution. 

If these results stand up to under further scrutiny, then we would need to reconcile this finding with our results in Table~\ref{tab:pooled_data}, which show that increased diversity (achieved by replacing some in-domain data with some data from another domain) degrades performance even for Places365. 
The simplest explanation is that the increased diversity here is much milder - we are simply changing how images are distributed over classes, not adding images from other datasets entirely. Our results indicate that this mild diversity is beneficial for pretraining, but too much diversity may render the contrastive pretraining task too easy, resulting in weaker features. 

\begin{table}[tb]
\centering
\scriptsize
\begin{tabular}{c|c|c|c|c}
Classes & Images / Class & ImageNet  & iNat21 & Places365\\ \hline 
200 & 500 & 0.509 & 0.314 & 0.390 \\
500 & 200 & 0.531 & 0.305 & 0.415 \\
1000 & 100 & 0.522 & 0.290 & 0.429 \\ \hline
\end{tabular}
\vspace{-6pt}
\caption{
\textbf{What is the effect of image diversity within a dataset?}
Linear evaluation results for self-supervised representations based on 100k ImageNet images distributed over different numbers of classes. 
}
\label{tab:data_diversity}
\end{table}

\section{Qualitative Examples}\label{sec:qualitative_examples}

\begin{figure*}
    \begin{subfigure}{\textwidth}
        \captionsetup{justification=centering}
        \includegraphics[width=0.19\textwidth]{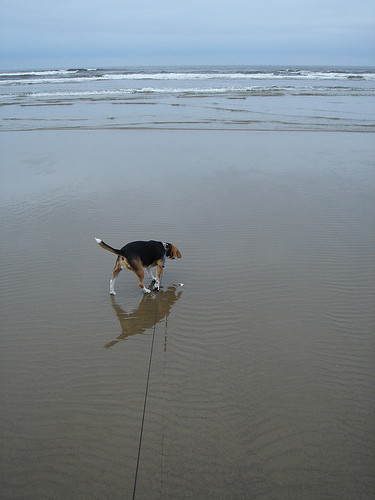}
        \includegraphics[width=0.19\textwidth]{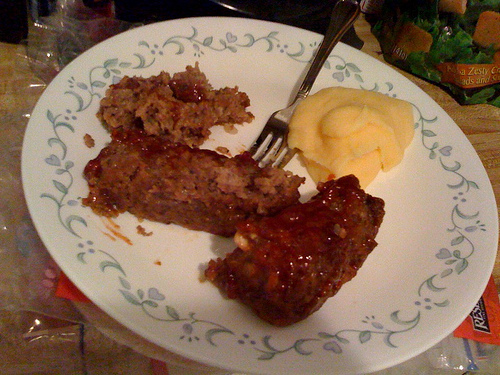}
        \includegraphics[width=0.19\textwidth]{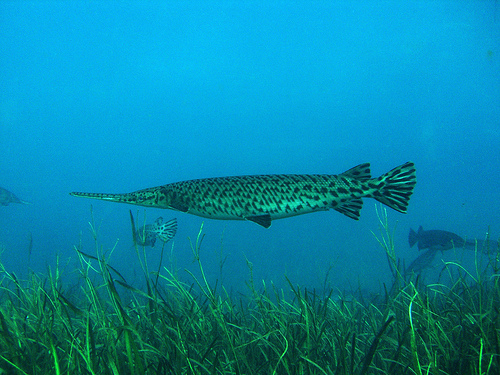}
        \includegraphics[width=0.19\textwidth]{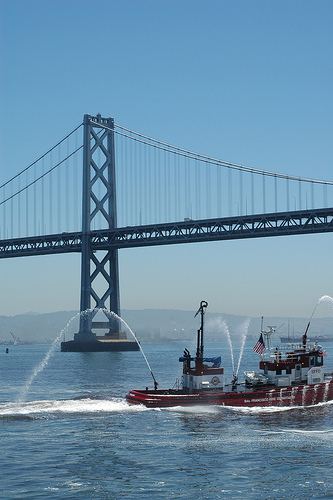}
        \includegraphics[width=0.19\textwidth]{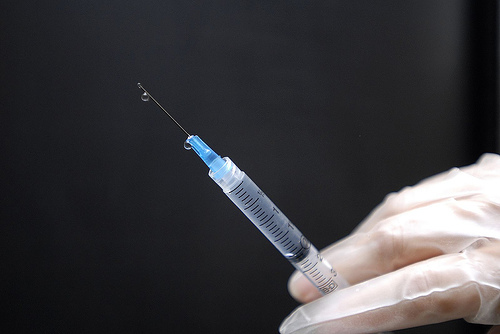}
        \caption{ImageNet}\vspace{5mm}
    \end{subfigure}
    \begin{subfigure}{\textwidth}
        \captionsetup{justification=centering}
        \includegraphics[width=0.19\textwidth]{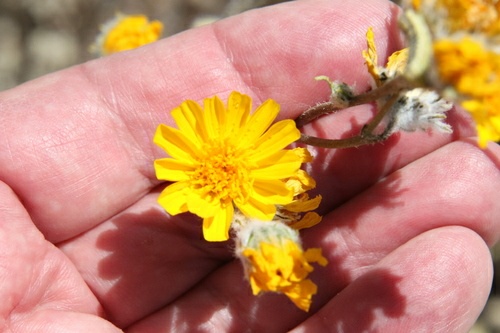}
        \includegraphics[width=0.19\textwidth]{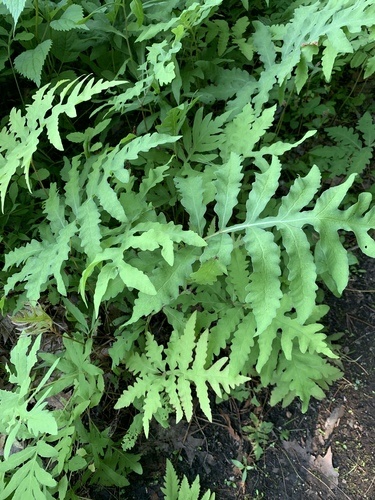}
        \includegraphics[width=0.19\textwidth]{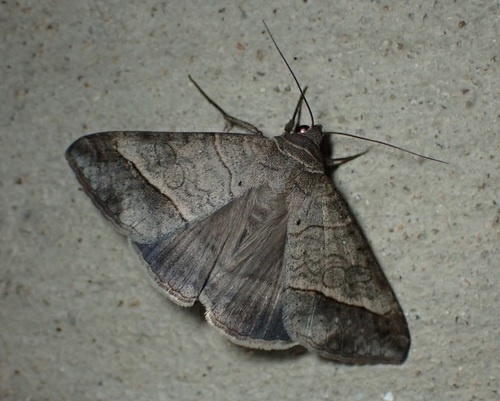}
        \includegraphics[width=0.19\textwidth]{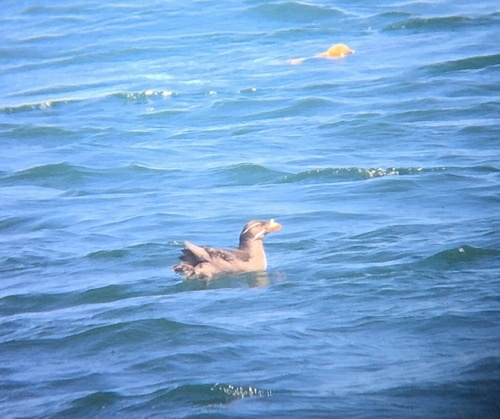}
        \includegraphics[width=0.19\textwidth]{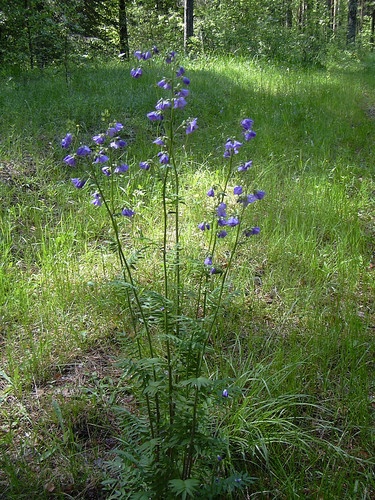}
        \caption{iNat21}\vspace{5mm}
    \end{subfigure}
    \begin{subfigure}{\textwidth}
        \captionsetup{justification=centering}
        \includegraphics[width=0.19\textwidth]{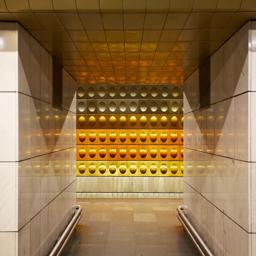}
        \includegraphics[width=0.19\textwidth]{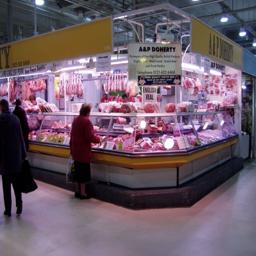}
        \includegraphics[width=0.19\textwidth]{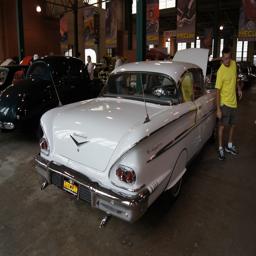}
        \includegraphics[width=0.19\textwidth]{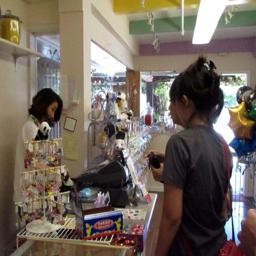}
        \includegraphics[width=0.19\textwidth]{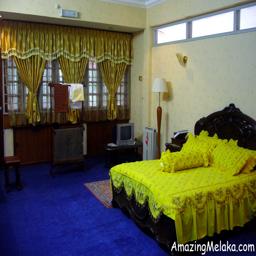}
        \caption{Places365}\vspace{5mm}
    \end{subfigure}
    \begin{subfigure}{\textwidth}
        \captionsetup{justification=centering}
        \includegraphics[width=0.19\textwidth]{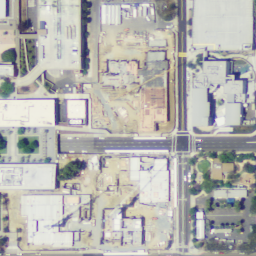}
        \includegraphics[width=0.19\textwidth]{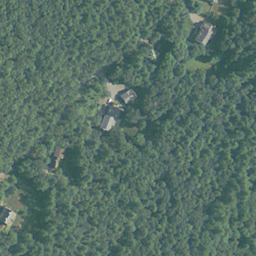}
        \includegraphics[width=0.19\textwidth]{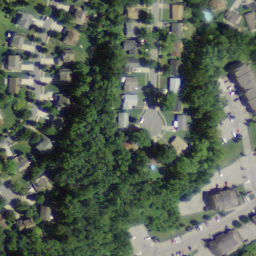}
        \includegraphics[width=0.19\textwidth]{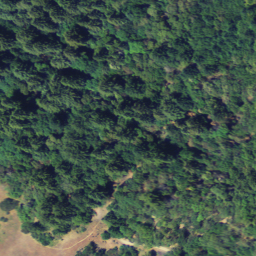}
        \includegraphics[width=0.19\textwidth]{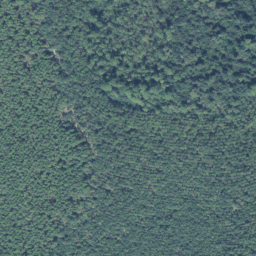}
        \caption{GLC20}
    \end{subfigure}
    \caption{
    \textbf{Examples from the datasets used.} We show five randomly selected images from each dataset: ImageNet (top row), iNat21 (second row), Places365 (third row), and GLC20 (bottom row). Note that all images in GLC20 and Places365 are 256$\times$256 pixels, while ImageNet and iNat21 have higher-resolution images and varying aspect ratios. 
    ``Places365-Standard" does have varying image resolutions, but we use ``Places365-Standard (small images)" which is an official variant that has been resized to 256$\times$256.
    }
    \label{fig:data_examples_supp}
\end{figure*}

\noindent
\textbf{Images from different domains.} In our paper we consider four datasets: ImageNet, iNat21, Places365, and GLC20. 
We illustrate their qualitative differences by showing some randomly chosen images from each dataset in Figure~\ref{fig:data_examples_supp}. 
By comparing the first row (ImageNet) with the second (iNat21) and third (Places365) rows, we can see that there are ImageNet images that are semantically similar to images from iNat21 (\eg the animals in the first and third images) and Places365 (\eg the bridge scene in the fourth image).
The images from GLC20 (bottom row) are quite distinct from the images from the other three datasets. 

\begin{figure*}
    \begin{subfigure}[t]{0.19\textwidth}
    \captionsetup{justification=centering}
    \includegraphics[width=\linewidth]{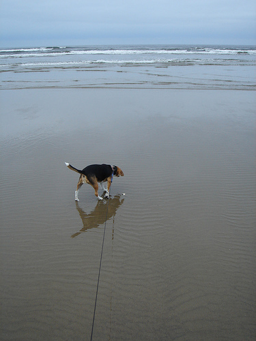}\\
    \includegraphics[width=\linewidth]{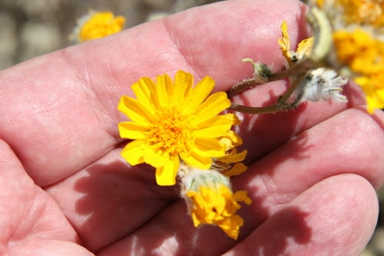}\\
    \includegraphics[width=\linewidth]{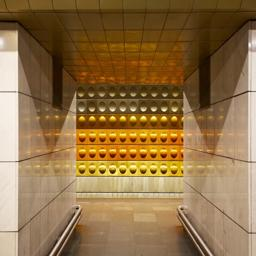}\\
    \includegraphics[width=\linewidth]{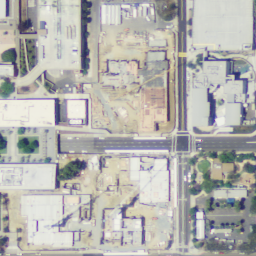}
    \caption{Resize}
    \end{subfigure}
    ~
    \begin{subfigure}[t]{0.19\textwidth}
    \captionsetup{justification=centering}
    \includegraphics[width=\linewidth]{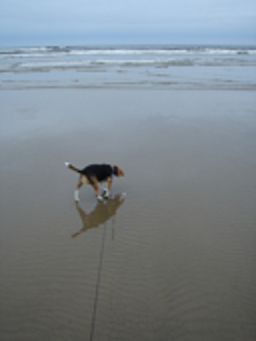}\\
    \includegraphics[width=\linewidth]{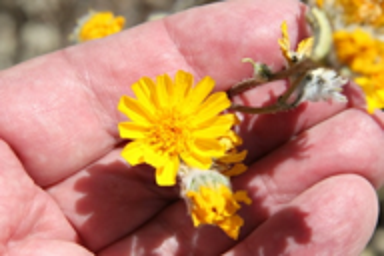}\\
    \includegraphics[width=\linewidth]{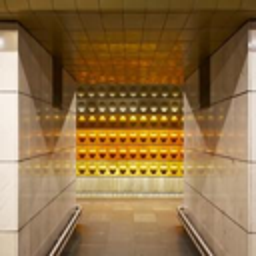}\\
    \includegraphics[width=\linewidth]{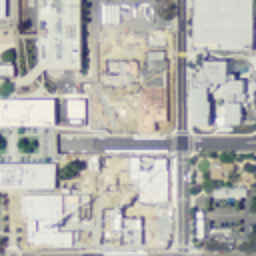}
    \caption{Resize \& Downsample (2x)}
    \end{subfigure}
    ~
    \begin{subfigure}[t]{0.19\textwidth}
    \captionsetup{justification=centering}
    \includegraphics[width=\linewidth]{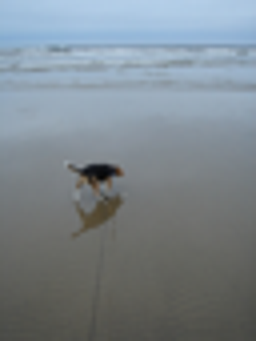}\\
    \includegraphics[width=\linewidth]{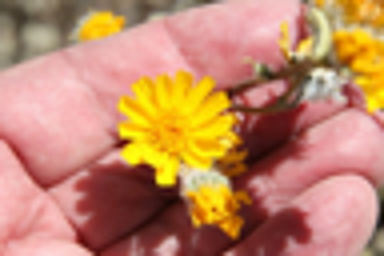}\\
    \includegraphics[width=\linewidth]{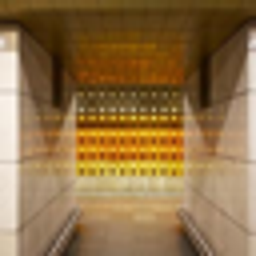}\\
    \includegraphics[width=\linewidth]{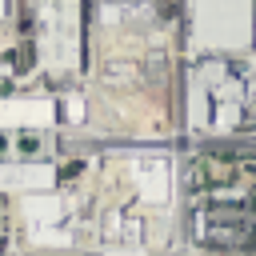}
    \caption{Resize \& Downsample (4x)}
    \end{subfigure}
    ~
    \begin{subfigure}[t]{0.19\textwidth}
    \captionsetup{justification=centering}
    \includegraphics[width=\linewidth]{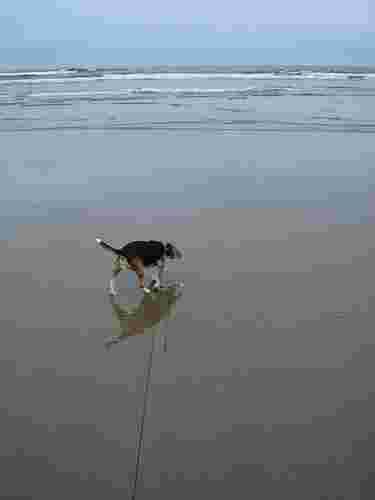}\\
    \includegraphics[width=\linewidth]{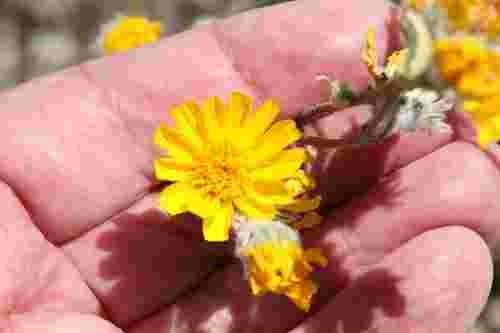}\\
    \includegraphics[width=\linewidth]{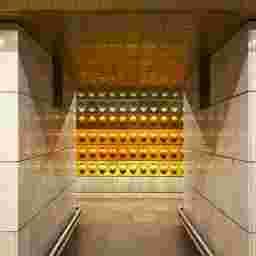}\\
    \includegraphics[width=\linewidth]{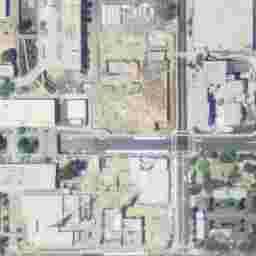}
    \caption{JPEG Compression}
    \end{subfigure}
    ~
    \begin{subfigure}[t]{0.19\textwidth}
    \captionsetup{justification=centering}
    \includegraphics[width=\linewidth]{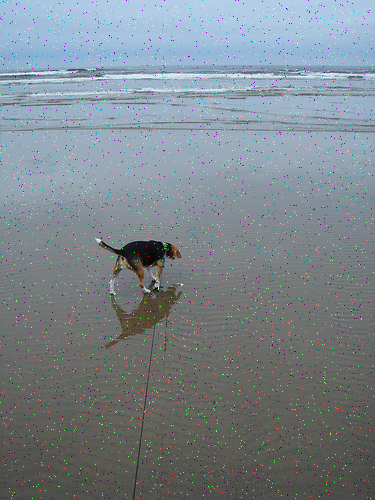}\\
    \includegraphics[width=\linewidth]{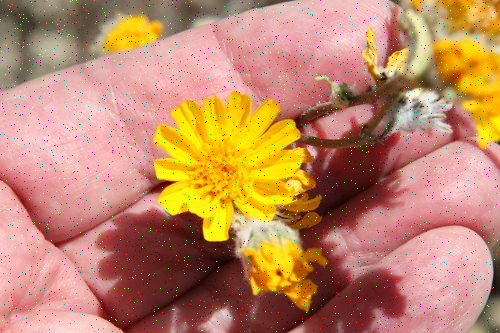}\\
    \includegraphics[width=\linewidth]{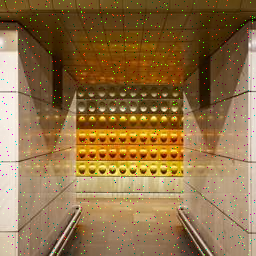}\\
    \includegraphics[width=\linewidth]{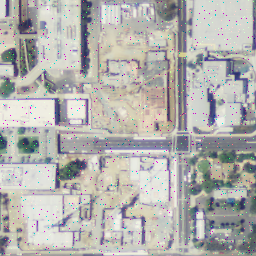}
    \caption{Salt \& Pepper}
    \end{subfigure}
    \caption{
    \textbf{Examples of corrupted images.}
    We show the effect of different image corruptions on one randomly chosen image from each dataset: ImageNet (top row), iNat21 (second row), Places365 (third row), and GLC20 (bottom row). 
    }
    \label{fig:corruption_examples_supp}
\end{figure*}

\noindent
\textbf{Corrupted images.} In Figure~\ref{fig:corruption_examples_supp} we show examples of the image corruptions we use in Figure~\ref{fig:image_corruption}. 
While all of these corruptions may seem subjectively mild, Figure~\ref{fig:image_corruption} shows that they can have a considerable impact on the quality of the learned representations.

\section{Implementation Details}\label{sec:implementation_details}

\subsection{Datasets}\label{sec:datasets}

\noindent
\textbf{iNat21.} The 2021 iNaturalist Challenge dataset (iNat21) is a fine-grained species classification dataset \cite{van2021benchmarking}, with  2.7M training images covering 10k species. Unlike prior iNaturalist datasets~\cite{van2018inaturalist}, iNat21 has an approximately balanced training set. 
The 100k official validation images are evenly sampled from each species, and we use it as our test set. 

\noindent
\textbf{GLC20.} GeoLifeCLEF 2020 \cite{cole2020geolifeclef} is a collection of remote sensing imagery, designed to facilitate research on location-based species prediction, while also serving as a land cover (LC) classification dataset. 
Each image is associated with a vector describing the distribution of land cover classes in a 256m$^2$ patch centered at that location.
For the purposes of this work, we binarize this vector (1 for any land cover class whose proportion is nonzero, 0 otherwise) and treat the task as multi-label classification. 
We only use the half of the dataset from the US, which means we have 1M training images covering 16 land cover classes. 
Throughout the paper, we refer to this subset of the GeoLifeCLEF 2020 dataset as GLC20. 
We use the official validation set as a test set, which has around 27k images that were held out in spatial blocks to mitigate the effects of spatial autocorrelation. 
Note that the labels for this dataset are noisy, so we are mainly interested in GLC20 as a pretraining set.

\subsection{Training hyperparameters}

\noindent
\textbf{SimCLR pretraining.}
Unless otherwise specified, we use the same settings as the ImageNet experiments in \cite{chen2020simple}.
One exception is that we omit Gaussian blur from the augmentation set since \cite{chen2020simple} found that it provides a relatively small benefit, around 1\% top-1 accuracy on ImageNet. 
Full details of the augmentations are given in Section \ref{sec:aug_implementation}.
We train with a batch size of 4096 for 1000 epochs and use 16 TPUs for training. 
We use the LARS optimizer \cite{you2017large} with a learning rate of $4.8$ (following $0.075 \times \text{batch size} / 256$), decayed on a cosine schedule \cite{loshchilov2016sgdr} with a $10$-epoch linear warmup and no restarts.
For small datasets (size 50k or smaller), we use a lower learning rate of $0.4$ (following $0.025 \times \text{batch size} / 256$) decayed on a cosine schedule.
Our projection head has two layers and an output dimension of $128$.
A temperature parameter of $\tau = 0.1$ is set for the contrastive loss.
Batch normalization statistics are synchronized across TPU cores with a decay parameter of $0.9$. 

\noindent
\textbf{MoCo pretraining.} We use the same settings as the ImageNet experiments in \cite{he2020momentum}, with the improvements noted in \cite{chen2020improved}. 
As in our SimCLR experiments, we train with a batch size of 1024 using 16 TPUs. 
For comparability, we use the same augmentation strategy as we do for SimCLR and train for 1000 epochs. 
Like \cite{chen2020simple} but unlike \cite{he2020momentum, chen2020improved}, we do not standardize images by subtracting per-channel means and dividing by per-channel standard deviations.

\noindent
\textbf{BYOL pretraining.} We use the same settings as the ImageNet experiments in \cite{grill2020bootstrap}. 
As in our SimCLR experiments, we train with a batch size of 4096 using 16 TPUs. 
For comparability, we use the same augmentation strategy as we do for SimCLR (which happens to be the default for \cite{grill2020bootstrap}) and train for 1000 epochs. 
Like \cite{chen2020simple} but unlike \cite{grill2020bootstrap}, we do not standardize images by subtracting per-channel means and dividing by per-channel standard deviations.

\noindent
\textbf{Linear supervised training.}
Linear classifiers are trained for 90 epochs using SGD with Nesterov momentum. We use a momentum of $0.9$, a batch size of 1024, and a learning rate of $0.4$, following the scaling rule $0.1 \times \text{batch size} / 256$. The learning rate follows a cosine decay schedule without linear warmup or restarts \cite{loshchilov2016sgdr}. 
Unless otherwise specified, we do not use weight decay / L2 regularization or data augmentation. 
We take a square center crop with edge length equal to 87.5\% of the short side of the image and resize to $224 \times 224$.
We use four Tesla V100 GPUs for training. 

\noindent
\textbf{End-to-end fine-tuning.}
We use the same settings as linear supervised training with the following exceptions.
We train using a smaller batch size of $512$ and a lower learning rate of $0.1$, following the learning rate scaling rule $0.05 \times \text{batch size} / 256$. 
To mitigate overfitting we use L2 regularization ($10^{-4}$) for the classifier head and data augmentation (random cropping and horizontal flips).
These augmentations use the same implementation as the cropping and flipping used for SimCLR pretraining. 

\noindent
\textbf{End-to-end supervised training from scratch.}
We use the same hyperparameters as end-to-end fine-tuning with the following exceptions.
We train for 90 epochs using a traditional piece-wise constant learning rate schedule where the initial learning rate of 0.1 is decayed by a factor of 10 at epochs 30 and 60. 
We also use L2 regularization of $10^{-4}$ throughout the network. 

\subsection{Taxonomies}

Three of our datasets are equipped with label taxonomies: ImageNet, iNat21, and Places365. 
We describe these taxonomies below.

\noindent
\textbf{ImageNet.} 
We use the WordNet~\cite{wordnet} label hierarchy for ImageNet. 
The finest labels are the standard ImageNet-1k class labels. 
To coarsen these labels, we start at the deepest level of the hierarchy and merge all leaf nodes at that level with their parents. 
This produces a new hierarchy, whose leaf nodes will now be used as categories. 
Each category set is named ``Depth $k$" where $k$ is the depth of the leaf node that is further from the root. 
We repeat this process until the leaf nodes merge with the root.

\noindent
\textbf{iNat21.} 
Since the categories in iNat21 are animal and plant species, the ``tree of life" serves as a natural taxonomy. 
The taxonomic levels are \emph{Species} (finest, 10k categories), \emph{Genus} (4884 categories), \emph{Family} (1103 categories), \emph{Order} (273 categories), \emph{Class} (51 categories), \emph{Phylum} (13 categories), and \emph{Kingdom} (coarsest, 3 categories).
For additional details see \cite{van2021benchmarking}.

\noindent
\textbf{Places365.}
Places 365 is equipped with a 3-tier hierarchy. The finest labels are the standard category labels for the dataset (``Depth 2"). 
These categories fall into 16 scene types which constitute the ``Depth 1'' level of the hierarchy. 
Examples include water, mountain, transportation, sports, industry, etc.
Then the ``Depth 0" level consists of a coarser grouping of these scene types into three categories: indoor, outdoor (natural), and outdoor (man-made). 

\subsection{Augmentations}\label{sec:aug_implementation}

In this paper we use three augmentation operations: random horizontal flipping, random cropping, and random color jittering. When training SimCLR, we use use all three augmentations. When fine-tuning we only use random horizontal flipping and random cropping as in \cite{chen2020simple}.
We do the same when training from scratch.
We do not use any data augmentation when training linear classifiers. 
For each of these operations we use the implementation from \cite{chen2020simple} with default settings. We give brief descriptions of each augmentation operation below.

\noindent
\textbf{Random horizontal flipping.} 
With probability $1/2$, flip the image over the vertical center line. 

\noindent
\textbf{Random cropping.} 
Randomly select a rectangular subset of the image covering between 8\% and 100\% of the whole image, with aspect ratio between $3/4$ and $4/3$.

\noindent
\textbf{Random color jitter.} 
Randomly perturb the brightness, contrast, saturation, and hue of the image according to a strength parameter $s$. See \cite{chen2020simple} for the exact implementation. We set the strength parameter to $s=1.0$.

\subsection{Corruptions}
In Section~\ref{sec:data_quality} of the main paper we investigate the impact of pretraining on artificially degraded images. 
Here we provide implementation details for each of the image corruption operations.

\noindent
\textbf{Resize.}
We resize the image so that the shorter side is 256 pixels long, but we preserve the aspect ratio. 
As described below, this corruption allows us to make comparisons which control for image size. 
Images are resized using the standard PIL \cite{pil} function \verb|PIL.Image.resize| with the default nearest-neighbor interpolation.

\noindent
\textbf{Resize and downsample.}
We first apply the ``Resize" corruption and then downsample by 2x or 4x before upsampling by the same factor. 
The initial resizing is important because some of our datasets have larger images than others and larger images are less affected by downsampling by a constant factor than their smaller counterparts.
Downsampling and upsampling is accomplished using \verb|PIL.Image.resize| with default settings, just like the ``Resize" corruption.

\noindent
\textbf{JPEG compression.}
We use the standard PIL function \texttt{PIL.Image.save} to perform JPEG compression. We set the \verb|quality| parameter to 10, which is low enough to cause significant visual artifacts.

\noindent
\textbf{Salt and pepper noise.} 
Each pixel in each channel is independently corrupted with probability $1/100$, and corrupted pixels are set to 0 (``pepper") or 1 (``salt") with equal probability.

\end{document}